\documentclass[10pt,twocolumn,letterpaper]{article}

\usepackage{style/cvpr}
\usepackage{times}
\usepackage{epsfig}
\usepackage{graphicx}
\usepackage{amsmath}
\usepackage{amssymb}

\usepackage{url}
\usepackage{multirow}
\usepackage{booktabs}
\usepackage{xcolor, soul}
\usepackage[ruled,lined,boxed]{algorithm2e}
\usepackage{algorithm2e}
\usepackage{verbatim}
\usepackage{xcolor}
\usepackage{adjustbox}
\usepackage{sidecap}
\sidecaptionvpos{figure}{c}
\usepackage{floatrow}
\usepackage[utf8x]{inputenc}
\usepackage{soul} 

\usepackage[breaklinks=true,bookmarks=false]{hyperref}

\cvprfinalcopy 

\def\cvprPaperID{7861} 

\ifcvprfinal\fi
\begin{document}

\title{Going Deeper with Lean Point Networks}

\author{Eric-Tuan Le$^{1}$
\and
Iasonas Kokkinos$^{1,2}$
\and
Niloy J. Mitra$^{1,3}$\smallbreak
\and
$^{1}$University College London \quad
$^{2}$Ariel AI \quad
$^{3}$Adobe Research
}

\newcommand{\pnpp}{PointNet++\xspace}
\newcommand{\mres}{mRes\xspace}
\newcommand{\mresx}{mResX\xspace}
\newcommand{\convp}{convPN\xspace}
\newcommand{\convpx}{convPNX\xspace}
\newcommand{\ngh}[1]{\mathcal{N}^{#1}}
\newcommand{\fvec}[1]{\mathbf{v}_{#1}}
\newcommand{\nghs}[2]{[#1,#2]}
\newcommand{\mlp}{\mathrm{MLP}}
\newcommand{\tensor}{T}
\newcommand{\cross}{\times}
\newcommand\blfootnote[1]{
  \begingroup
  \renewcommand\thefootnote{}\footnote{#1}%
  \addtocounter{footnote}{-1}
  \endgroup
}
\newcommand{\name}{\textsc{PointNetLite}\xspace}
\newcommand{\todo}[1]{{\color{red}#1}}
\newcommand{\refeq}[1]{Eq.~\ref{#1}}
\newcommand{\reffig}[1]{Fig.~\ref{#1}}
\newcommand{\refsec}[1]{Sec.~\ref{#1}}
\newcommand{\fix}{\marginpar{FIX}}
\newcommand{\new}{\marginpar{NEW}}

\newcommand\mycommfont[1]{\footnotesize\ttfamily\textcolor{blue}{#1}}
\SetCommentSty{mycommfont}
\newcommand\mynlfont[1]{\scriptsize\sffamily{#1}}
\SetNlSty{mynlfont}{}{}
\LinesNumbered
\SetAlFnt{\footnotesize}
\newcommand{\myfuncsty}[1]{\textcolor{black}{\textbf{\texttt{#1}}}}
\SetFuncSty{myfuncsty}

\newcommand{\beginsupplement}{%
    \setcounter{section}{0}
    \renewcommand{\thesection}{S\arabic{section}}%
    \setcounter{table}{0}
    \renewcommand{\thetable}{S\arabic{table}}%
    \setcounter{figure}{0}
    \renewcommand{\thefigure}{S\arabic{figure}}%
}

\maketitle
\thispagestyle{empty}

\newcommand{\mycomment}[1]{}
\newcommand{\ikchange}[1]{{\color{red} {#1}}}

\begin{abstract}
In this work we introduce Lean Point Networks (LPNs) to train deeper and more accurate point processing networks by relying on three novel point  processing blocks that improve memory consumption, inference time, and accuracy: a \textit{convolution-type  block}  for point sets that blends neighborhood information in a memory-efficient manner; a \textit{crosslink block} that efficiently shares information across low- and high-resolution processing branches; and a \textit{multi-resolution point cloud processing block} for faster diffusion of information. By combining these blocks, we design wider and deeper point-based architectures.
%
We report systematic accuracy and memory consumption improvements
on multiple publicly available segmentation tasks
by using our generic modules as drop-in replacements for the blocks of  multiple  architectures (PointNet++, DGCNN, SpiderNet, PointCNN). Code is publicly available at  \href{https://geometry.cs.ucl.ac.uk/projects/2020/deepleanpn/}{geometry.cs.ucl.ac.uk/projects/2020/deepleanpn/}.  
\end{abstract}

\section{Introduction}
\label{sec:introduction}

\begin{figure}[bh!]
    \centering
        \centering
        \includegraphics[width=0.9\linewidth]{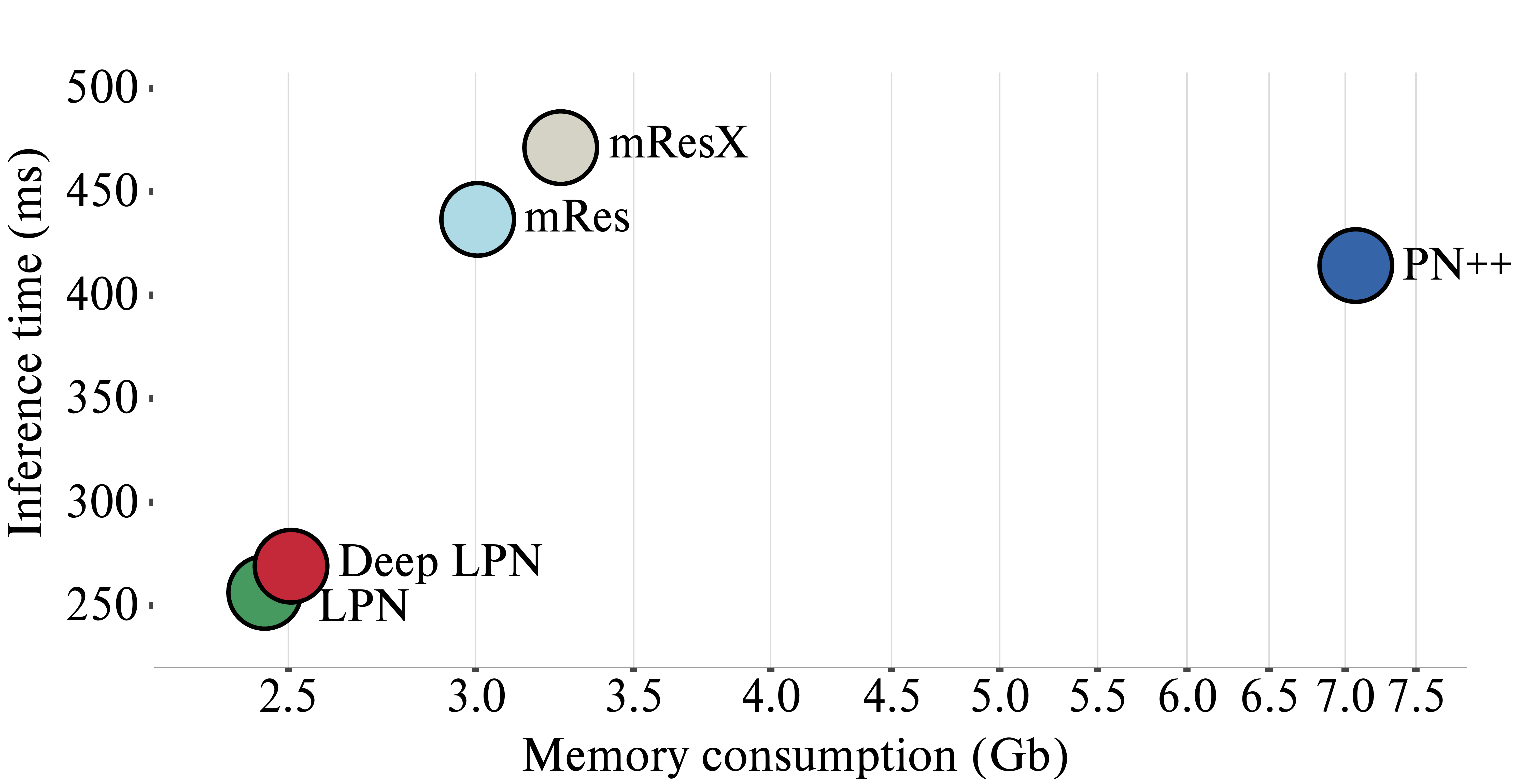}%
        \vspace{5pt}
        \includegraphics[width=0.8\linewidth]{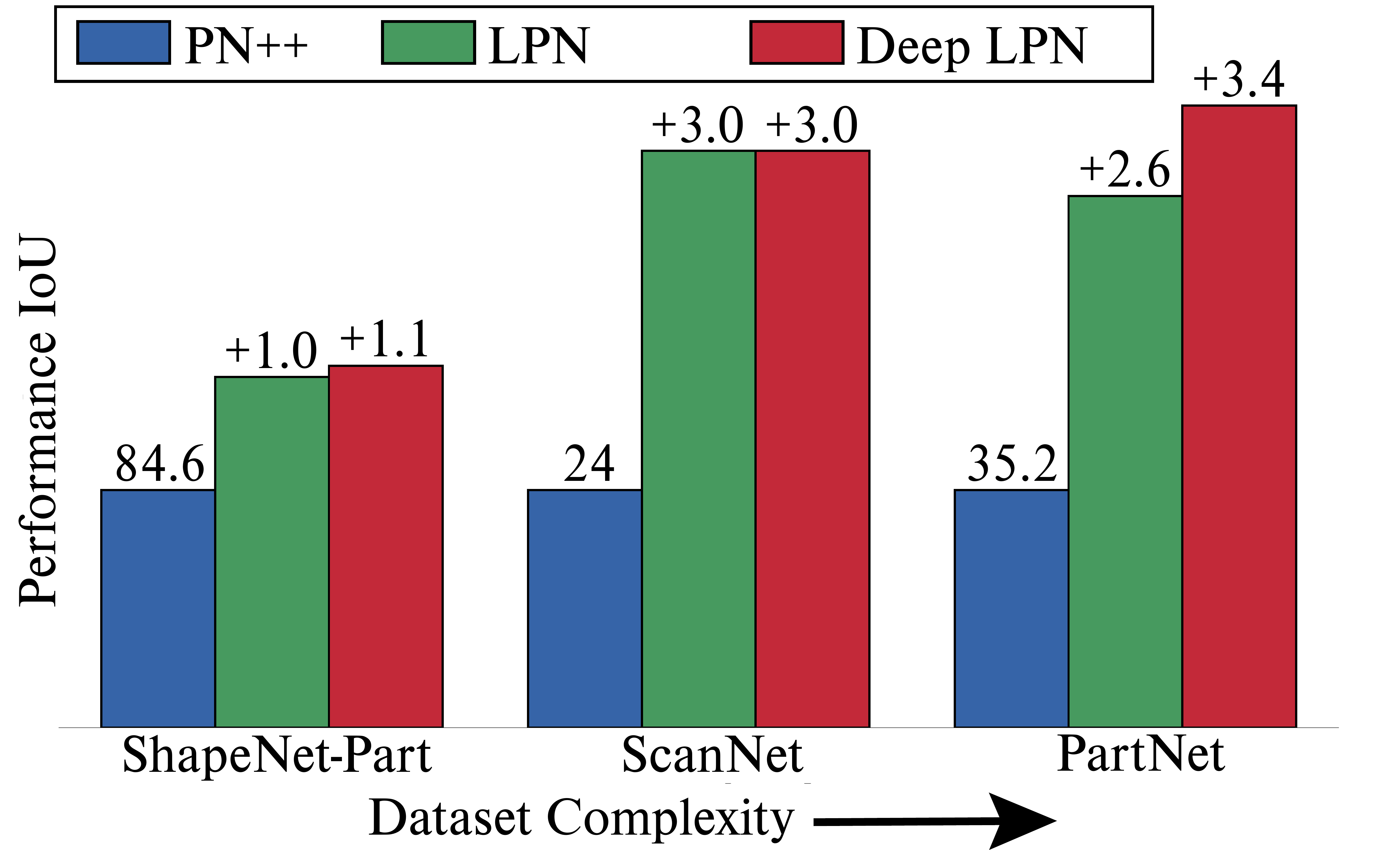}
        \vspace{5pt}
        \includegraphics[width=0.9\linewidth]{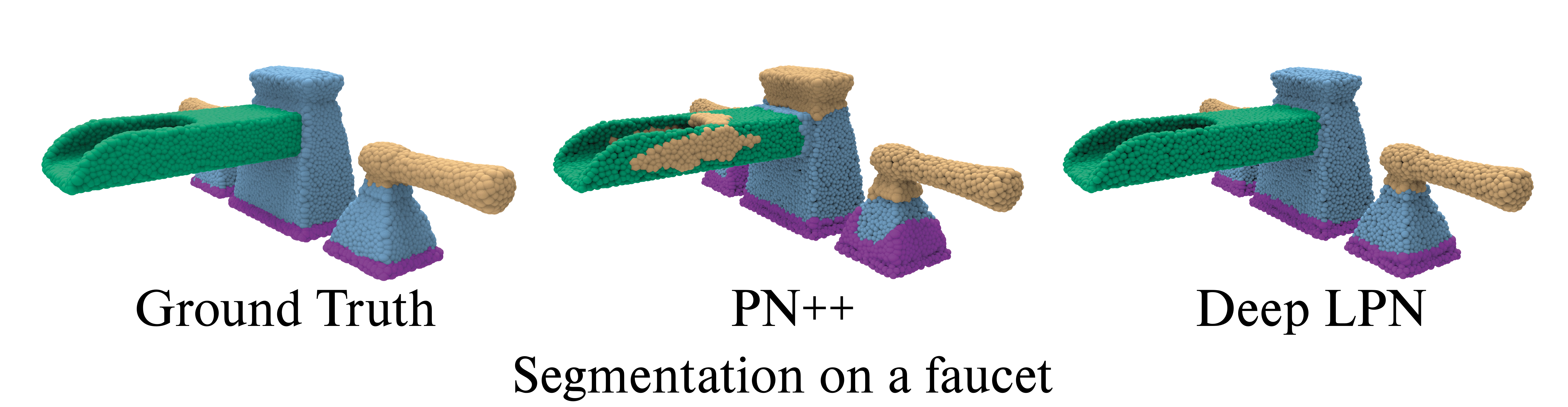}
    \caption{Lean Point Networks (LPNs) can achieve higher point cloud segmentation accuracy while operating at substantially lower memory and inference time. (Top) Memory footprint and inference speed of LPN variants introduced in this work
    compared to the  PointNet++ (PN++) baseline. (Middle) Improvements in accuracy for three segmentation benchmarks of increasing complexity. On the --most complex-- PartNet dataset our deep network outperforms the shallow PointNet++ baseline by $3.4\%$, yielding a $9.7\%$ relative increase. (Bottom) Part Segmentation by PointNet++ and Deep LPN.}
    \label{fig:iouVScomplexity}
\end{figure}

Geometry processing has started profiting from applying deep learning to graphics and 3D shape analysis~\cite{QiEtAl:Pointnet++:NIPS:2017,WangEtAl:DGCNN:arxiv:2018,dai2017scannet,matterport17}, delivering networks that guarantee desirable properties of point cloud processing, such as permutation-invariance and quantization-free representation \cite{su15mvcnn,Wang-2017-OCNN,Wang-2018-AOCNN}.
Despite these advances, the memory requirements of a majority of the point processing architectures still impede  breakthroughs similar to those made in computer vision. 

Directly working with unstructured representations of 3D data (i.e., not residing on a spatial grid), necessitates re-inventing, for geometry processing, the functionality of basic image processing blocks, such as convolution operations, information exchange pathways, and multi-resolution data flow. 
Taking a Lagrangian perspective, in order to avoid quantization or multi-view methods, the pioneering PointNet architecture~\cite{QiEtAl:Pointnet:CVPR:2017} demonstrated how to directly work on point cloud datasets by first lifting individual points to higher dimensional features (via a shared MLP) and then performing permutation-invariant local pooling. 
PointNet++~\cite{QiEtAl:Pointnet++:NIPS:2017}, by considering local point patches,  groups information within a neighborhood based on Euclidean distance and then applies PointNet to the individual groups. This design choice requires explicitly duplicating local memory information among neighboring points, and potentially compromises performance as the networks go deeper. 

In particular, the memory and computational demands of  point network blocks (e.g., PointNet++ and many follow-up architectures) can affect both training speed and, more crucially, inference time. 
One of the main bottlenecks for such point networks is their memory-intensive nature, as detailed in \refsec{pointnet}. Specifically, the PointNet++ architecture and its variants replicate point neighborhood information, letting every node carry in its feature vector information about all of its neighborhood. This results in significant memory overhead, and limits the number of layers, features and  feature compositions one can compute. 

%

In this work, we enhance such point processing networks that replicate local neighborhood information by introducing a set of modules that improve  memory footprint and accuracy, without compromising on inference speed. We call the resulting architectures \textit{Lean Point Networks}, to highlight their lightweight memory budget. 
We build on the decreased memory budget to go deeper with point networks. As  has been witnessed repeatedly in the image domain \cite{HeZRS15,densenet,wrnet}, we show that going deep also increases the prediction accuracy of point networks.

We start in \refsec{conv} by replacing the grouping operation used in point cloud processing networks with a low-memory alternative that is the point cloud processing counterpart of efficient image processing implementations of convolution. The resulting \emph{`point convolution block'} -- defined slightly differently from convolution used in classical signal processing -- is 67\% more memory-efficient and 41\% faster than its PointNet++ counterpart, while exhibiting favorable training properties due to more effective mixing of information across neighborhoods.

We then turn  in \refsec{optim} to improving the information flow across layers and scales within point networks through three techniques: a \emph{multi-resolution} variant for multi-scale network which still delivers the multi-scale context but at a reduced memory and computational cost, \emph{residual links}, and a new \emph{cross-link block} that broadcasts multi-scale information across the network branches.
By combining these blocks, we are able to successfully train deeper point networks that allow us to leverage upon larger  datasets. 

In  \refsec{sec:experiments} we  validate our method on the ShapeNet-Part, ScanNet and PartNet segmentation benchmarks, reporting systematic improvements over the PointNet++ baseline. As shown in \reffig{fig:iouVScomplexity}, when combined these contributions deliver multifold reductions in memory consumption while improving performance, allowing us in a second stage to train increasingly wide and deep networks. On PartNet, the most complex dataset, our deep architecture achieves a 9.7\% relative increase in IoU while decreasing memory footprint by 57\% and inference time by 47\%. 

Having ablated our design choices on the PointNet++ baseline, in \refsec{subsec:results_diffArchitectures} we turn to confirming the generic nature of our blocks. We extend the scope of our experiments to three additional networks, (i)~DGCNN~\cite{WangEtAl:DGCNN:arxiv:2018}, (ii) SpiderCNN~\cite{xu2018spidercnn}, and (iii) PointCNN~\cite{li2018pointcnn} 
and report systematic improvements in memory efficiency and performance. 



\section{Related Work}
\label{sec:relatedWork}
 
{\bf Learning in Point Clouds.} Learning-based approaches have recently attracted significant attention in the context of Geometric Data Analysis, with
several methods proposed specifically to handle point cloud data, including PointNet~\cite{QiEtAl:Pointnet:CVPR:2017} and several extensions such as PointNet++~\cite{QiEtAl:Pointnet++:NIPS:2017} and Dynamic Graph CNNs~\cite{WangEtAl:DGCNN:arxiv:2018} for shape
segmentation and classification, PCPNet~\cite{GuerreroEtAl:PCPNet:EG:2018} for normal and curvature
estimation, P2P-Net~\cite{Yin:2018:PBP:3197517.3201288} and PU-Net~\cite{yu2018pu} for cross-domain point cloud transformation.  
Alternatively, kernel-based methods~\cite{Atzmon:2018:PCN:3197517.3201301,hermosilla2018mccnn,DBLP:journals/corr/abs-1904-07601,DBLP:journals/corr/abs-1904-08889} have also been proposed with impressive performance results. 
Although many alternatives to PointNet have been proposed \cite{su2018splatnet, li2018sonet, li2018pointcnn, hermosilla2018mccnn, zaheer2017deepsets} to achieve higher performance, the simplicity and effectiveness of PointNet and its extension PointNet++ make it popular for many other tasks~\cite{yu2018ecnet}.

Taking PointNet++ as our starting point, our work facilitates the transfer of network design techniques developed in computer vision to point cloud processing. In particular, significant accuracy improvements have been obtained with respect to the original AlexNet network  \cite{KrizhevskyNIPS2013} by engineering the scale of the filtering operations \cite{ZeilerF14, simonyan2014very},  the structure of the computational blocks \cite{SzegedyLJSRAEVR14,DBLP:journals/corr/XieGDTH16}, and the network's width and depth \cite{HeZRS15,wrnet}. A catalyst for experimenting with a larger space of network architecture, however, is the reduction of  memory consumption - this motivated us to design lean alternatives to point processing networks. Notably, \cite{zhang-shellnet-iccv19} introduce a new operator to improve point cloud network efficiency, but only focus on increasing the convergence speed by tuning the receptive field.  \cite{li2019deepgcns} has investigated how residual/dense connections and dilated convolution could help mitigate vanishing gradient observed for deep graph convolution networks but without solving memory limitations, \cite{hermosilla2018mccnn} use Monte Carlo estimators to estimate local convolution kernels. 
By contrast our work explicitly tackles the memory problem with the objective of training deeper/wider networks and shows that there are clear improvements over strong baselines. 

\mycomment{
\begin{figure*}[ht!]
    \centering
    \includegraphics[width=\textwidth]{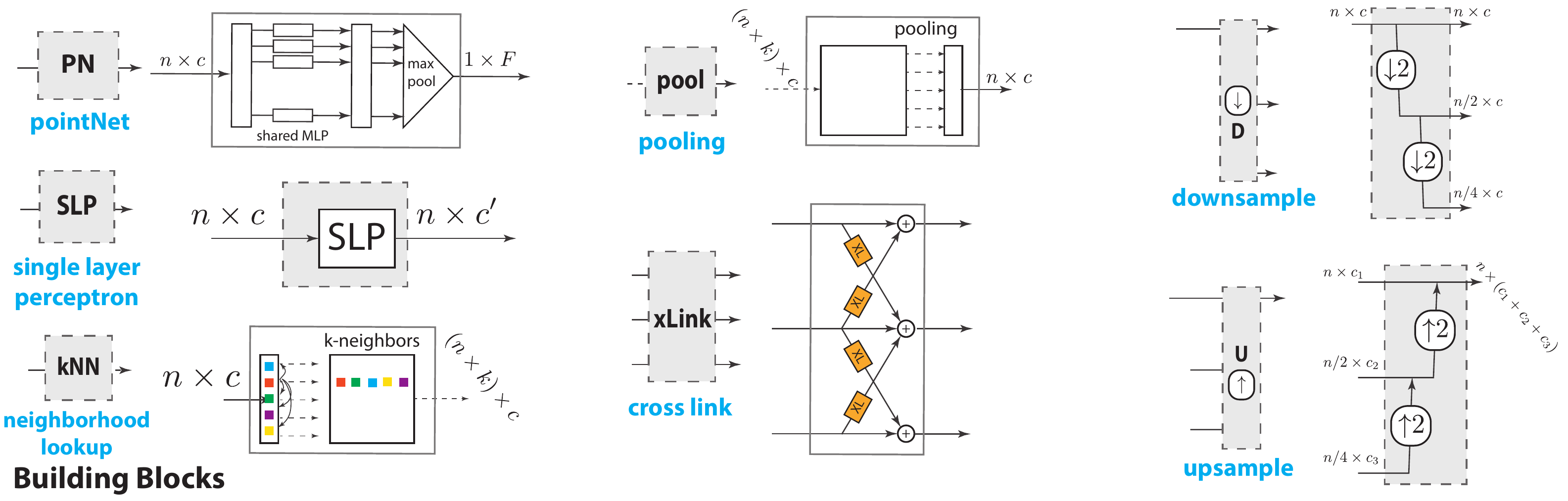}
    \caption{
     Elementary building blocks for point processing. Apart from standard neighborhood lookup, pooling and SLP layers, we introduce cross-link layers across scales, and propose multi-resolution up/down sampling blocks for point processing. PointNet module combines a stack of shared SLP (forming an MLP) to lift individual points \iffalse to higher dimensional features \fi and then performs permutation-invariant local pooling.}
    \label{fig:pipeline_blocks}
\end{figure*}
}

{\bf Memory-Efficient Networks.}
The  memory complexity of the standard back-propagation implementation grows linearly with network's depth as backprop requires retaining in memory all of the intermediate activations computed during the forward pass, since they are required for the gradient computation in the backward pass. 

Several methods bypass this problem by trading off speed with memory. 
Checkpointing techniques \cite{ChenXZG16,NIPS2016_6221}
use anchor points to free up intermediate computation results, and re-compute them in the backward pass. This is 1.5x slower during training, since one performs effectively two forward passes rather than just one. More importantly, applying this technique is easy for chain-structured graphs, e.g., recursive networks  \cite{NIPS2016_6221} but is not as easy for general Directed Acyclic Graphs, such as U-Nets, or multi-scale networks like \pnpp. One needs to manually identify the graph components, making it cumbersome to experiment with diverse variations of architectures.

Reversible Residual Networks (RevNets) \cite{GomezRUG17} limit the computational block to come in  a particular, invertible form of residual network. This is also 1.5x slower during training, but alleviates the need for anchor points altogether. Unfortunately, it is unclear what is the  point cloud counterpart of invertible blocks.

We propose generic blocks to reduce the memory footprint inspired from multi-resolution processing and efficient implementations of the convolution operation in computer vision. As we show in \refsec{subsec:results_diffArchitectures}, our blocks can be used as drop-in replacements in generic point processing architectures (PointNet++, DGCNN, SpiderNet, PointCNN) without any additional network design effort. 
\section{Method}
\label{sec:method}
We start with a brief introduction of the PointNet++ network, which serves as an example point network baseline. We then introduce our modules and explain how they decrease memory footprint and improve information flow.  

\subsection{PointNet and PointNet++ Architectures}
\label{pointnet}
\begin{figure*}[t!]
    \includegraphics[width=\textwidth]{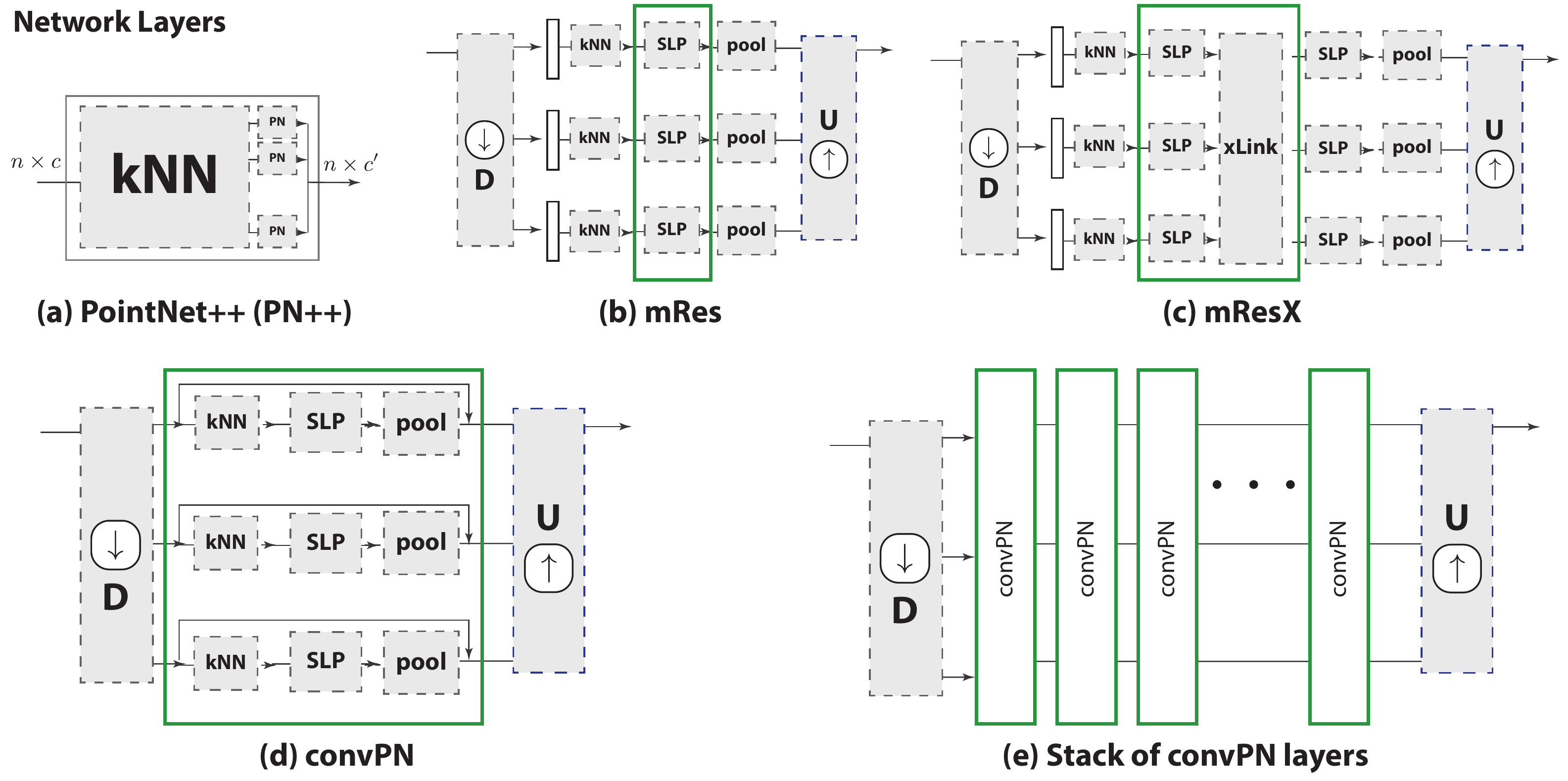}
    \caption{
     The standard PN++ layer in (a) amounts to the composition of a k-Nearest Neighbor (kNN)-based lookup and a PointNet element. In (b) we propose to combine parallel \pnpp blocks in a multi-resolution architecture (D and U stand for down- and up-sampling operations), using multiple Single Layer Perceptrons (SLPs) and in (c) allow information to flow across  branches of different resolutions through a cross-link element ('xLink').
     In (d) we propose to turn the lookup-SLP-pooling cascade  into a low-memory counterpart by removing the kNN elements from memory once computed; we also introduce residual links, improving the gradient flow. In (e) we stack the green box in (d) to grow in depth and build our deep architecture. 
    }
    \label{fig:pipeline_modules}
\end{figure*}

PointNet++~\cite{QiEtAl:Pointnet++:NIPS:2017} has been proposed as a method of augmenting the  basic PointNet architecture with a grouping operation.
 As shown in \reffig{fig:pipeline_modules}(a), grouping is implemented through a `neighborhood lookup',  where each
point $\mathbf{p}_i$ looks up its $k$-nearest neighbors and stacks them to get a point set, say $P_{N_k}^i$.
If each point comes with a $D$-dimensional feature vector, the result of this process is a tensor 
$T=[\begin{array}{cccc} \fvec{} &\fvec{\nghs{.}{1}}&\ldots&\fvec{\nghs{.}{K}}\end{array}]$ of size $N\cross D \cross (K+1)$.
Within the PointNet modules, every vector of this matrix is processed separately by a Multi-Layer-Perceptron that implements a function 
$\mlp:  R^D \to R^{D^{'}}$, while at a later point a max-pooling operation over the $K$ neighbors of every point delivers a slim, $N\cross D^{'}$ matrix.

When training a network every layer constructs and retains  a matrix like $T$ in memory, so that it can be used in the backward pass to update 
the MLP parameters, and send gradients to earlier layers. While demonstrated to be extremely effective, this design of \pnpp has two main shortcomings: 
  first, because of explicitly carrying around $k$-nearest neighbor information for each point, the network layers are memory intensive; 
  and second, being reliant on PointNet, it also delays transmission of global information until the last, max-pooling stage where the results of decoupled MLPs are combined. 
  Many subsequent variants of PointNet++ suffer from similar memory and information flow limitations. As we describe now, these shortcomings are
  alleviated, by our convolution-type point processing layer.

\mycomment{
\begin{algorithm}[b!]
\DontPrintSemicolon
\KwData{Input features tensor $\mathcal{T}_f$ ($N\times R^D$), input spatial tensor $\mathcal{T}_s$ ($N\times R^3$) and indices of each point's neighborhood for lookup operation $\mathcal{L}$ ($N\times K$)}
\KwResult{Output feature tensor $\mathcal{T}_f^o$ ($N\times R^{D^{'}}$)}
\SetKwFunction{IndexLookup}{IndexLookup}
\SetKwFunction{SLP}{SLP}
\SetKwFunction{MaxPooling}{MaxPooling}
\SetKwFunction{FreeMemory}{FreeMemory}
\Begin{
\tcc*[h]{Lifting each point/feature to $R^{D^{'}}$}\;
$\mathcal{T}_{f'}\longleftarrow\SLP_f(\mathcal{T}_f)$\;
$\mathcal{T}_{s'}\longleftarrow\SLP_s(\mathcal{T}_s)$\;

\tcc*[h]{Neighborhood features $(N\times R^{D^{'}} \rightarrow N\times R^{D^{'}} \times (K+1))$}\;
$\mathcal{T}_{f'}^K\longleftarrow\IndexLookup(\mathcal{T}_{f'},\mathcal{T}_{s'},\mathcal{L})$\;

\tcc*[h]{Neighborhood pooling $(N\times R^{D^{'}}\times (K+1) \rightarrow N\times R^{D^{'}})$}\;
$\mathcal{T}_{f^{'}}^o\longleftarrow\MaxPooling(\mathcal{T}_{f^{'}}^{K})$\;
$\FreeMemory(\mathcal{T}_{s^{'}},\mathcal{T}_{f^{'}},\mathcal{T}_{f^{'}}^K)$\;
\Return $\mathcal{T}_{f^{'}}^o$
}
\caption{Low-memory grouping - Forward pass}
\label{algo:convpn_forward}
\end{algorithm}
}

\begin{figure*}
    \includegraphics[width=1\columnwidth]{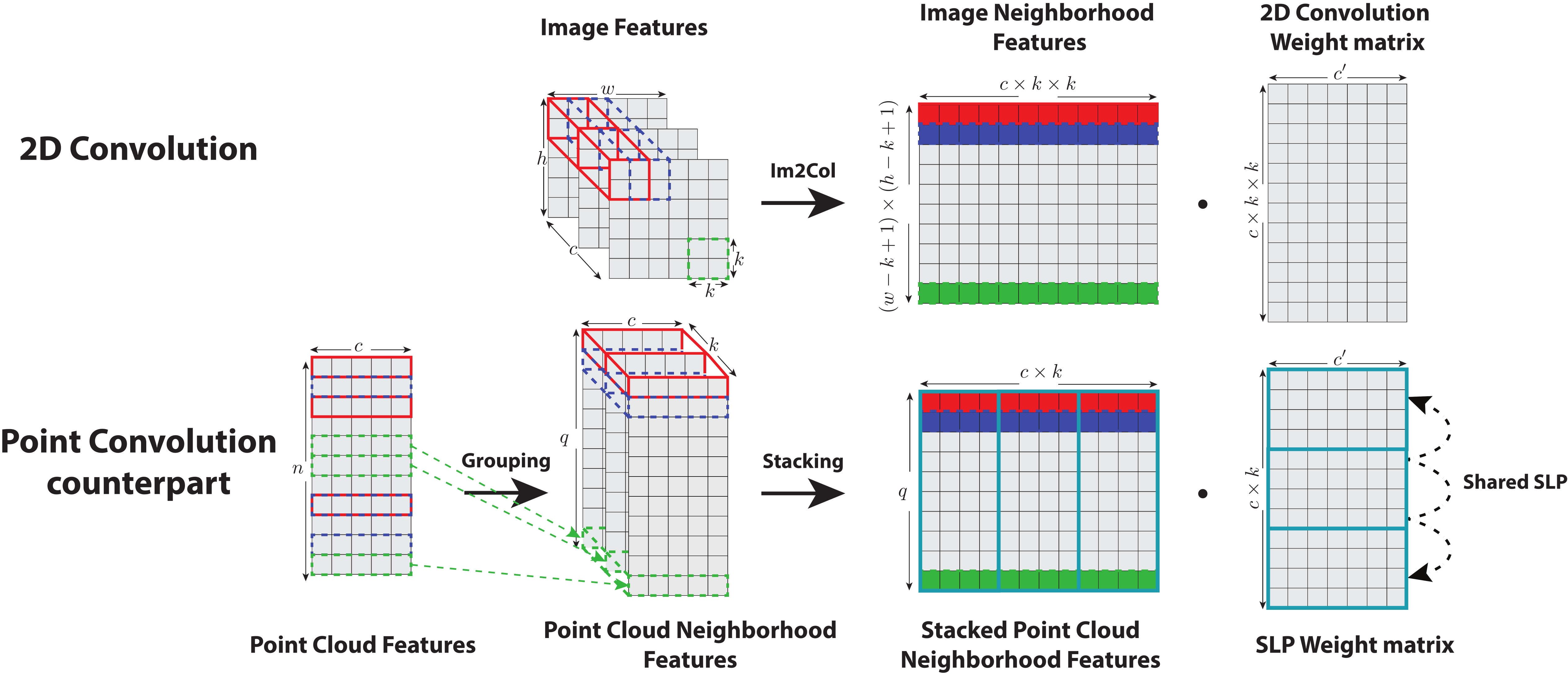}
    \centering
    \caption{Analogy between 2D convolution and its point cloud counterpart. In both cases, the layer operates in two steps: (i) neighborhood exposure and (ii) matrix multiplication. 2D image convolution amounts to forming a $K^2$ tensor in memory when performing $K\times K$ filtering and then implementing a convolution as matrix multiplication. \mycomment{In point clouds the nearest neighbor information provides us with the counterpart to the $K\times K$ neighborhood.}}
    \label{fig:2d_convolution}
\end{figure*}

\begin{figure*}
    \includegraphics[width=1\columnwidth]{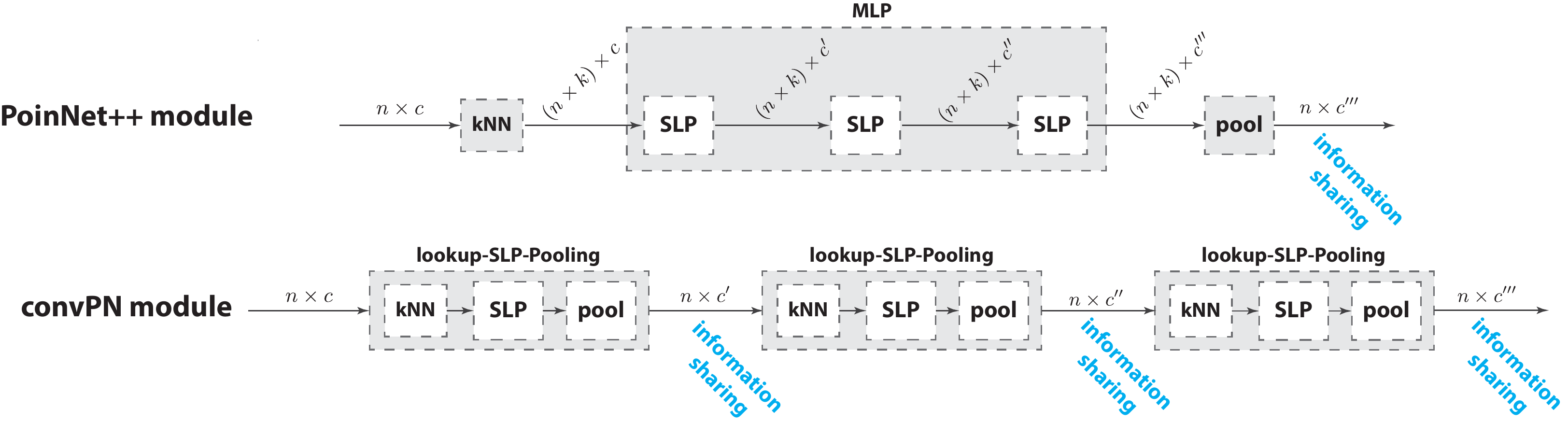}
    \centering
    \caption{Comparison of a PointNet++ module with our convPN module. The convPN module replaces the MLP with its pooling layer by a sequence of SLP-Pooling modules which has two benefits (i) memory savings as the layer activations are saved only through the pooled features and (ii) better information flow as it increases the frequency at which neighbors share information. \mycomment{This design requires however to do neighborhood lookups back again in-between the layers}}
    \label{fig:exposing_neighbours}
\end{figure*}

\subsection{convPN: a convolution-type PointNet layer}
\label{conv}

\mycomment{
As described above, the existing \pnpp grouping operation exposes the neighborhood of any point $i$ by concatenating all of its $K$ neighboring $D$-dimensional vectors  $\fvec{\nghs{i}{k}}$ to form a tensor with the neighborhood lookup module:
$T=[\begin{array}{cccc} \fvec{} &\fvec{\nghs{.}{1}}&\ldots&\fvec{\nghs{.}{K}}\end{array}]$ of size $N\cross D \cross (K+1)$. Within the PointNet modules, every vector of this matrix is processed separately by a Multi-Layer-Perceptron that implements a function $\mlp:  R^D \to R^{D^{'}}$, while at a later point a max-pooling operation over the $K$ neighbors of every point delivers a slim, $N\cross D^{'}$ matrix.
When training a network every layer constructs and retains such a matrix in memory, so that it can be used in the backward pass to update the MLP parameters, and send gradients to earlier layers. 
}
We propose  a novel  convolution-type PointNet layer (convPN), that is inspired from efficient implementations of  convolution. Standard convolution operates in two steps: (i) neighborhood exposure and (ii) matrix multiplication. Our convPN block follows similar steps, as shown in \reffig{fig:2d_convolution}, but the weight matrix blocks treating different neighbors are tied, so as to ensure permutation invariance. 

In more detail, standard 2D image convolution amounts to forming a $K^2$ tensor in memory when performing $K\times K$ filtering and then implementing a convolution as matrix multiplication. This amounts to the \texttt{im2col} operation used to implement convolutions with General Matrix-Matrix Multiplication (GEMM) \cite{Jia14}. In point clouds the nearest neighbor information provides us with the counterpart to the $K\times K$ neighborhood. 

Based on this observation we propose to use the strategy used in memory-efficient implementations of image convolutions for deep learning: we free the memory from a layer as soon as the forward pass computes its output, rather than maintaining the matrix in memory. In the backward pass we reconstruct the matrix {\emph{on the fly}} from the outputs of the previous layer. We perform the required gradient computations and then return the GPU memory resources.

As shown in \reffig{fig:exposing_neighbours}, we gain further efficiency by replacing the MLPs of \pnpp by a sequence of SLP-Pooling modules. This allows us to further reduce memory consumption, saving the layer activations only through the pooled features while at the same time increasing the frequency at which neighbors share information.




As detailed in the Supplemental Material,  a careful implementation of our convolution-type architecture shortens, on average, the time spent for the forward pass and the backward pass by 41\% and 68\%,  respectively, while resulting in a drastic reduction in memory consumption.

\mycomment{
\begin{algorithm}
\DontPrintSemicolon
\KwData{Input features tensor $\mathcal{T}_f$ ($N\times R^D$), input spatial tensor $\mathcal{T}_s$ ($N\times R^3$), gradient of the output $\mathcal{G}_{out}$ and indices of each point's neighborhood for lookup $\mathcal{L}$ ($N\times K$)}
\KwResult{Gradient of the input $\mathcal{G}_{in}$ and gradient of the weights $\mathcal{G}_{w}$}
\SetKwFunction{IndexLookup}{IndexLookup}
\SetKwFunction{InverseIndexLookup}{InverseIndexLookup}
\SetKwFunction{BackwardMaxPooling}{BackwardMaxPooling}
\SetKwFunction{BackwardSLP}{BackwardSLP}
\SetKwFunction{FreeMemory}{FreeMemory}
\Begin{
\tcc*[h]{Gradient Max Pooling $(N\times R^{D^{'}} \rightarrow N\times R^{D^{'}}\times (K+1))$}\;
$\mathcal{G}_{out}^{mp}\longleftarrow\BackwardMaxPooling(\mathcal{G}_{out})$\;
\tcc*[h]{Flattening features $(N\times R^{D^{'}}\times (K+1) \rightarrow N\times R^{D^{'}})$}\;
$\mathcal{G}_{out}^{fl}\longleftarrow\InverseIndexLookup(\mathcal{G}_{out}^{mp},\mathcal{L})$ \;
\tcc*[h]{Gradient wrt. input/weight}\;
$\mathcal{G}_{w}, \mathcal{G}_{in}\longleftarrow\BackwardSLP(\mathcal{T}_f,\mathcal{T}_s,\mathcal{G}_{out}^{fl})$\;
$\FreeMemory(\mathcal{T}_f,\mathcal{T}_s,\mathcal{G}_{out},\mathcal{G}_{out}^{mp},\mathcal{G}_{out}^{fl})$\;
\Return $(\mathcal{G}_{in},\mathcal{G}_{w})$\;
}
\caption{Low-memory grouping - Backward pass}
\label{algo:convpn_backward}
\end{algorithm}
}

\begin{figure}[t!]
    \includegraphics[width=1\columnwidth]{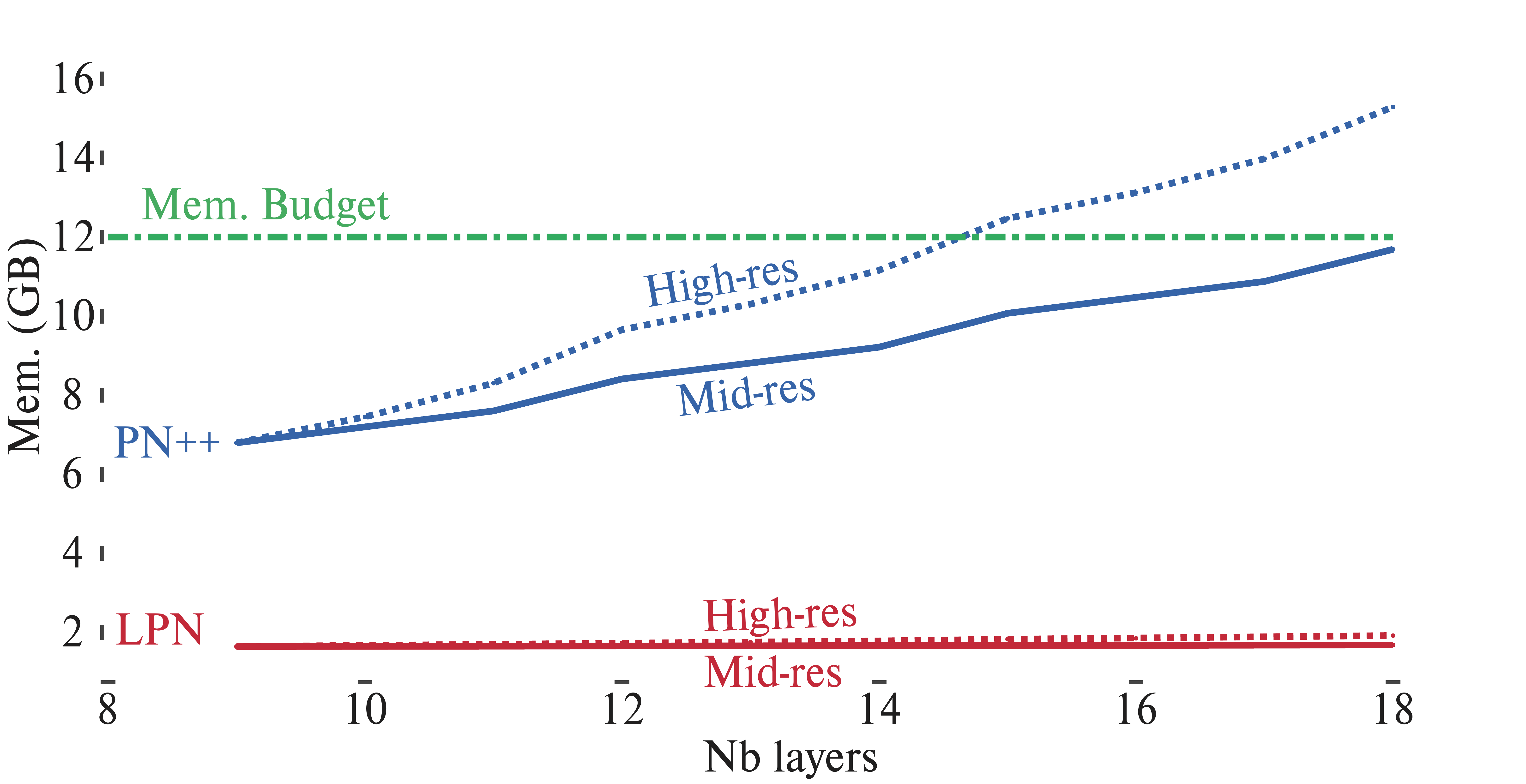}
    \caption{Evolution of memory consumption as the number of layers increases for PointNet++ and LPN (convolution block counterpart) on ShapeNet-Part. Doubling the number of layers for LPN results only in an increase in memory by +2.3\% and +16.8\% for mid- and high- resolution respectively, which favorably compares to the +72\% and +125\% increases for PointNet++.}
    \label{fig:deepVSmemory}
\end{figure}

For a network with $L$ layers, the  memory consumption of the baseline PointNet++ layer grows as $L\cross( N \cross D \cross K)$, while in our case memory consumption grows as  
$L\cross( N \cross D ) + (N \cross D \cross K)$, where the  term, $L\cross( N \cross D )$ accounts for the memory required to store the layer activations, while the second term $N \cross D \cross K$ is the per-layer memory consumption of a single neighborhood convolution layer. 
As $L$ grows larger, this results in a $K$-fold  drop, shown on \reffig{fig:deepVSmemory}. This reduction opens up the possibility of learning much deeper networks, since memory demands now grow substantially more slowly in depth. With minor, dataset-dependent, fluctuations, the memory footprint of our convolution type architecture is on average 67\% lower than the \mbox{\pnpp} baseline, while doubling the number of layers comes with a memory overhead  of 2.7\%.

\subsection{Improving Information Flow}
\label{optim}
We now turn to methods for efficient information propagation through point networks. As has been repeatedly shown in computer vision, this can  drastically impact  the behavior of the network during training. Our experiments indicate that this is also the case for point processing.

{\bf (a) Multi-Resolution vs Multi-Scale Processing.} 

Shape features can benefit from both local, fine-grained information and global, semantic-level context; their  fusion can easily boost the discriminative power of the resulting features. 
For this we  propose to extract neighborhoods of fixed size in downsampled versions of the original point cloud.
In the coordinates of the original point cloud this amounts to increasing the effective grouping area, but it now comes with a much smaller memory budget. We observe a 58\% decrease in memory footprint on average on the three tested datasets. Please refer to in Supplemental for an illustration of the difference between both types of processing.

{\bf (b) Residual Links.} We use the standard Residual Network architecture~\cite{HeZRS15}, which helps to train deep networks reliably.
Residual networks change  the network's connectivity to improve gradient  flow  during training: identity connections provide early network layers with access to undistorted versions of the loss gradient, effectively mitigating the vanishing gradient problem. As our results in \refsec{sec:experiments} show,  this allows us to train deeper networks.

\begin{figure}[ht!]
    \centering
    \includegraphics[width=0.5\textwidth]{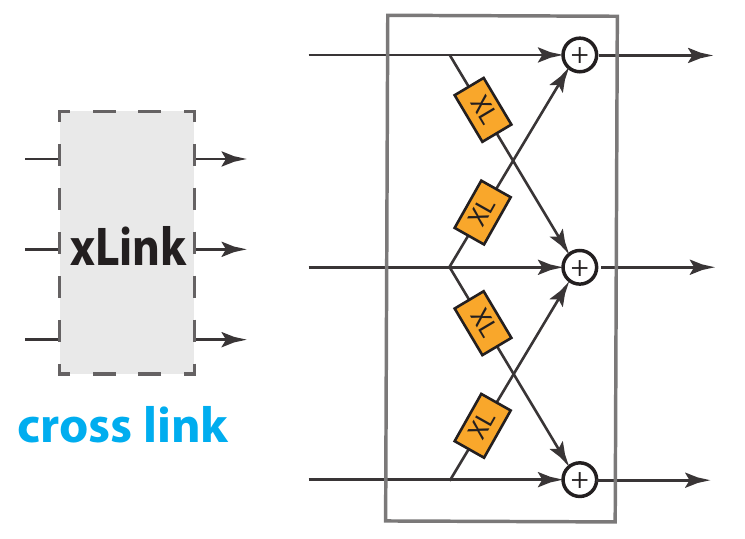}
    \caption{
     Cross-link module to connect across resolutions.}
    \label{fig:xlink_block}
\end{figure}
{\bf (c) Cross Links.} Further, we use  Cross-Resolution Links to better propagate information in the network during training. We draw inspiration from the Multi-Grid Networks~\cite{DBLP:journals/corr/KeMY16}, Multiresolution Tree Networks~\cite{gadelha2018multiresolution}, Hypercolumns~\cite{DBLP:journals/corr/HariharanAGM14a};  and allow layers that reside in different resolution branches to communicate with each other, thereby exchanging low-, mid-, and high-resolution information throughout the network processing, rather than fusing multi-resolution information at the end of each block. 

Cross-links broadcast information across resolutions as shown in \reffig{fig:xlink_block}. Note that unlike \cite{gadelha2018multiresolution}, an MLP transforms the output of one branch to the right output dimensionality so that it can be combined with the output of another branch. Each resolution branch can focus on its own representation and the MLPs will be in charge of making the translation between them. Taking in particular the case of a high-resolution branch communicating its outputs to a mid-resolution branch, we have $N\times D^H$ feature vectors at the output of a lookup-SLP-pooling block cascade, which need to be communicated to the $N/2 \times D^M$
vectors of the mid-resolution branch. We first downsample the points, going from $N$ to $N/2$ points, and then use an  MLP that transforms the vectors to the target dimensionality. Conversely, when going from low- to higher dimensions we first transform the points to  the right dimensionality and then upsample them. We have experimented with both concatenating and summing multi-resolution features and have observed that summation behaves systematically better in terms of both training speed and test performance.
\section{Evaluation}
\label{sec:experiments}
We start by defining our tasks and metrics and then turn to validating our two main contributions to model accuracy, namely better network training through improved network flow in \refsec{subsec:information_flow}, and going deeper through memory-efficient processing in \refsec{subsec:going_deeper}. We then turn to validating the merit of our modules when used in tandem with a broad array of state-of-the-art architectures in \refsec{subsec:results_diffArchitectures}, and finally provide a thorough ablation of the impact of our contributions on aspects complementary to accuracy, namely parameter size, memory, and efficiency in \refsec{subsec:ablation_study}.

{\bf Dataset and evaluation measures.}
Our modules can easily be applied to any point cloud related tasks, such as classification, however, we focus here on evaluating our modules on the point cloud segmentation task on three different datasets as it is a more challenging task.
The datasets consist of either 3D CAD models or real-world scans. We quantify the complexity of each dataset based on (i)~the number of training samples, (ii)~the homogeneity of the samples and (iii)~the granularity of the segmentation task. Note that a network trained on a bigger and diverse dataset would be less prone to overfitting - as such we can draw more informative conclusions from more complex datasets. We order the datasets by increasing complexity: ShapeNet-Part~\cite{shapenet2015}, ScanNet~\cite{dai2017scannet} and PartNet \cite{DBLP:journals/corr/abs-1812-02713} for fine-grained segmentation. By its size (24,506 samples) and its granularity (251 labeled parts), PartNet is the most complex dataset.

To have a fair comparison (memory, speed, accuracy), we re-implemented all the models in Pytorch and consistently compared the vanilla network architectures and our memory-efficient version. We report the performance of  networks using their last saved checkpoint (i.e. when training converges), instead of the common (but clearly flawed) practice of using the checkpoint that yields best performance on the test set. These two factors can lead to small differences  from the originally reported performances.

We use two different metrics to report the Intersection over Union (IoU): (i) the mean Intersection over Union ($\text{mIoU}$) and (ii) the part Intersection over Union ($\text{pIoU}$). Please refer to Supplemental for further details.

\subsection{Effect of improved information flow}
\label{subsec:information_flow}
We report the performance of our variations for PointNet++ on the Shapenet-Part, ScanNet and PartNet datasets (Table \ref{tab:accuracyDatasets}).
Our lean and deep architectures can  be easily deployed on large and complex datasets. Hence, for PartNet, we choose to train on the full dataset all at once on a segmentation task across the 17 classes instead of having to train a separate network for each category as in \cite{DBLP:journals/corr/abs-1812-02713}.

Our architectures substantially improve the memory efficiency of the PointNet++ baseline while also delivering an increase in performance for more complex datasets (see \reffig{fig:iouVScomplexity}). Indeed, as the data complexity grows, having efficient information flow has a larger influence on the network performance. On PartNet, the spread between our architectures and the vanilla \pnpp becomes significantly high: our multiresolution (\mres) network increases relative performance by +5.7\% over \pnpp and this gain reaches +6.5\% with cross-links (\mresx). Our convolution-type network (LPN)   outperforms other architectures when dataset complexity increases (+3.4\% on ScanNet and +7.4\% on PartNet) by more efficiently mixing  information across neighbours.

\begin{table}[t!]
  \centering
  \caption{Performance of our modules compared to PointNet++ baseline. The impact of our modules becomes most prominent as the dataset complexity grows. On PartNet our Deep LPN network  increases pIoU by 9.7\% over  \pnpp, outperforming its shallow counterpart by +2.1\%.}
  \resizebox{\textwidth}{!}{
    \begin{tabular}{r|c|cc|c}
\cline{2-5}    \multicolumn{1}{c}{} & ShapeNet-Part & \multicolumn{2}{c|}{ScanNet} & PartNet \\
\multicolumn{1}{c}{} & (13,998 samp.) & \multicolumn{2}{c|}{(1,201 samp.)} & (17,119 samp.) \\
\cline{2-5}    \multicolumn{1}{c}{} & mIoU (\%) & Vox. Acc. (\%) & pIoU (\%) & pIoU (\%) \\
    \hline
    PN++  & 84.60 (+0.0\%) & 80.5 (+0.0\%) & 24 (+0.0\%) & 35.2 (+0.0\%) \\
    \hline
    mRes  & 85.47 (+1.0\%) & 79.4 (-1.4\%) & 22 (-8.3\%) & 37.2 (+5.7\%) \\
    \hline
    mResX & 85.42 (+1.0\%) & 79.5 (-1.2\%) & 22 (-8.3\%) & 37.5 (+6.5\%) \\
    \hline
    LPN & 85.65 (+1.2\%) & \textbf{83.2 (+3.4\%)}    & \textbf{27 (+12.5\%)} & 37.8 (+7.4\%) \\
    \hline
    Deep LPN & \textbf{85.66 (+1.3\%)} & 82.2 (+2.1\%) & \textbf{27 (+12.5\%)} & \textbf{38.6 (+9.7\%)} \\
    \hline
    \end{tabular}}
  \label{tab:accuracyDatasets}%
\end{table}%

\begin{table*}[t!]
\small
  \centering
  \caption{Performance of our deepConPN network compared to Deep GCN (ResGCN-28) and related methods on S3DIS based on a 6-fold validation process. The difference in performance observed on each class can be explained by the different approaches networks have for point convolution. Our deep network clearly outperforms PointNet++ baseline by a spread of +6.8\% for mIoU. We achieve similar performance compared to Deep GCN while relying on a weaker baseline (PointNet++ against DGCNN)}
  \setlength{\tabcolsep}{4.2pt}
    \begin{tabular}{r|cc|ccccccccccccc}
    \hline
    Method & OA    & mIOU  & ceiling & floor & wall  & beam  & column & window & door  & table  & chair & sofa  & bookcase & board & clutter \\
    \hline
    MS+CU & 79.2  & 47.8  & 88.6  & \textbf{95.8}  & 67.3  & 36.9  & 24.9  & 48.6  & 52.3  & 51.9  & 45.1  & 10.6  & 36.8  & 24.7  & 37.5 \\
    G+RCU & 81.1  & 49.7  & 90.3  & 92.1  & 67.9  & 44.7  & 24.2  & 52.3  & 51.2  & 58.1  & 47.4  & 6.9   & 39.0  & 30.0  & 41.9 \\
    3DRNN+CF & \textbf{86.9}  & 56.3  & 92.9  & 93.8  & 73.1  & 42.5  & 25.9  & 47.6  & 59.2  & 60.4  & 66.7  & 24.8  & \textbf{57.0}  & 36.7  & 51.6 \\
    \hline
    DGCNN & 84.1  & 56.1  & -     & -     & -     & -     & -     & -     & -     & -     & -     & -     & -     & -     & - \\
    ResGCN-28 & 85.9  & \textbf{60.0}  & \textbf{93.1}  & 95.3  & \textbf{78.2}  & 33.9  & \textbf{37.4}  & \textbf{56.1}  & \textbf{68.2}  & 64.9  & 61.0  & \textbf{34.6}  & 51.5  & 51.1  & 54.4 \\
    \hline
    PointNet & 78.5  & 47.6  & 88.0  & 88.7  & 69.3  & 42.4  & 23.1  & 47.5  & 51.6  & 54.1  & 42.0  & 9.6  & 38.2  & 29.4  & 35.2 \\
    PointNet++ & -     & 53.2  & 90.2  & 91.7  & 73.1  & 42.7  & 21.2  & 49.7  & 42.3  & 62.7  & 59.0  & 19.6  & 45.8  & 48.2  & 45.6 \\
    Deep LPN & 85.7 & \textbf{60.0} & 91.0 & 95.6      & 76.1      & \textbf{50.3}      & 25.9       & 55.1      &  56.8     &  \textbf{66.3}     &  \textbf{74.3}     & 25.8      & 54.0      & \textbf{52.3}      &  \textbf{55.3}       \\
    \hline
    \end{tabular}%
  \label{tab:benchmark_deepgcn}%
  \vspace{-.1in}
\end{table*}%

\begin{figure*}[t!]
    \includegraphics[width=\columnwidth]{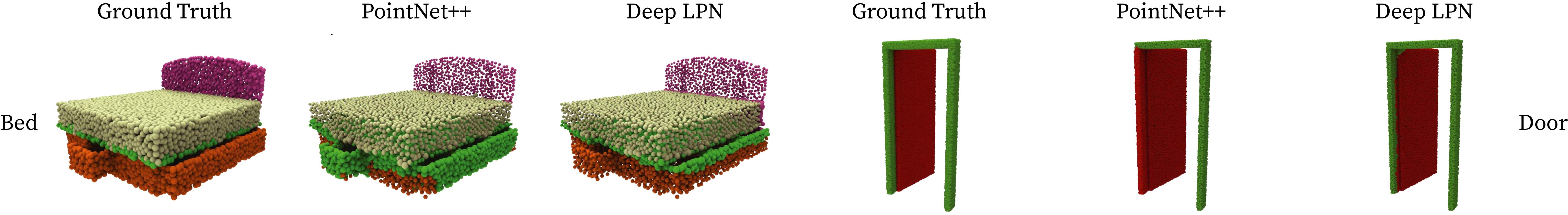}
    \centering
    \caption{Segmentation prediction for both PointNet++ and  Deep LPN networks compared to the ground truth. While PointNet++ struggles to detect accurately the boundaries between different parts, ours performs a much finer segmentation in those frontier areas.}
    \label{fig:rendering_segmentation}
\end{figure*}

\subsection{Improvement of accuracy by going deeper}
\label{subsec:going_deeper}
The  memory savings introduced in Sec.~\ref{conv} give the opportunity to design deeper networks. Naively increasing network depth can harm performance~\cite{HeZRS15}. Instead, we use residual connections to improve convergence for our deep network. The architecture, detailed  in the Supplemental material, consists in doubling the number of layers in the encoding part. While keeping the impact on efficiency very small (+6.3\% on inference time on average and +3.6\% on memory consumption at most compared to the shallow LPN), the performance improved (see Table~\ref{tab:accuracyDatasets}). On PartNet, this margin reaches +2.1\% over the shallow LPN and +9.7\% over the vanilla \pnpp. Note the low growth of memory as a function of depth, shown in  \reffig{fig:deepVSmemory}. As shown in \reffig{fig:rendering_segmentation}, having a deeper network improves the segmentation quality at the boundaries between parts. In contrast,  naively increasing the number of encoding layers from 9 to 15 in PointNet++ leads only to a small increase in performance IoU from 84.60\% to 84.66\%.

In Table~\ref{tab:benchmark_deepgcn} we compare against Deep GCN~\cite{li2019deepgcns} in terms of the overall accuracy and the mean IoU on the S3DIS dataset~\cite{armeni_cvpr16} by following the same 6-fold evaluation process. We  attain similar performance to Deep GCN, while relying on our generic memory-efficient network blocks and while based on a weaker baseline compared to Deep GCN (i.e., DGCNN). Moreover, as Deep GCN is not designed to tackle the efficiency issue in Point Networks, our network wins on all counts and successfully reduces both the memory (more than 74\%) and the speed (-48\% and -89\% for the inference and the backward speed respectively). As we show in the following section, these blocks come with the advantage of being applicable to many other point processing networks.

\subsection{Evaluation on more architectures}
\label{subsec:results_diffArchitectures}

We have introduced building blocks for point processing networks based on two key ideas, (i) a memory efficient convolution and (ii) improved information flow. Our blocks make it really efficient to capture, process and diffuse information in a point neighbourhood. Diffusing information across neighborhood is the main behavior that most networks, if not all, share.
We validate the generality of the proposed modular blocks in the context of other point-based learning setups, as  shown in Table \ref{tab:performance_architectures}. Each of our macro-blocks can be stacked together, extended into a deeper block by duplicating the green boxes (see Figure~\ref{fig:pipeline_modules}) or even be modified by changing one of its components by another.
We test our framework on three additional networks among the recent approaches, (i) Dynamic Graph CNN \cite{WangEtAl:DGCNN:arxiv:2018}, (ii) PointCNN \cite{li2018pointcnn} and (iii) SpiderCNN \cite{xu2018spidercnn}. These networks involve a diverse set of point convolution approaches; these experiments  allow us to assess the generic nature  of our modular blocks, and their value as drop-in replacements for existing layers.

All three of the networks make extensive use of memory which is a bottleneck to depth. We implant our modules directly in the original networks, making, when needed, some approximations from the initial architecture (see Supplemental). We report the performance of each network with our lean counterpart on two metrics: (i) memory footprint and (ii) accuracy in Table \ref{tab:performance_architectures}. Our lean counterparts consistently improve both the accuracy (from +0.4\% up to +8.0\%) and the memory consumption (from -19\% up to -69\%).

\begin{table*}[htbp]
  \footnotesize
  \centering
  \caption{Performance of our blocks on three different architectures (DGCNN, PointCNN and SpiderCNN) on three datasets using two different metrics: (i) memory consumption in Gb and (ii) performance in \% (mIoU for ShapeNet-Part, Vox. Acc. for ScanNet and pIoU for PartNet). Our lean counterparts improve significantly both the performance (up to +8.0\%) and the memory consumption (up to -69\%).}
   \setlength{\tabcolsep}{2.7pt}
    \begin{tabular}{c|c|ccc|ccc|ccc|ccc|}
\cline{3-11}    \multicolumn{1}{r}{} &       & \multicolumn{3}{c|}{DGCNN} & \multicolumn{3}{c|}{PointCNN} & \multicolumn{3}{c|}{SCNN} \\
\cline{3-11}    \multicolumn{1}{c}{} &       & ShapeNet-P & ScanNet & PartNet & ShapeNet-P & ScanNet & PartNet & ShapeNet-P & ScanNet & PartNet \\
    \hline
    \multirow{2}[1]{*}{Mem.} & Vanilla & 2.62 (+0\%) & 7.03 (+0\%) & 9.50 (+0\%) & 4.54 (+0\%) & 5.18 (+0\%) & 6.83 (+0\%) & 1.09 (+0\%) & 4.33 (+0\%) & 5.21 (+0\%) \\
          & Lean & \textbf{0.81 (-69\%)} & \textbf{3.99 (-43\%)} & \textbf{5.77 (-39\%)} & \textbf{1.98 (-56\%)} & \textbf{3.93 (-24\%)} & \textbf{5.55 (-19\%)} & \textbf{0.79 (-28\%)} & \textbf{3.25 (-25\%)} & \textbf{3.33 (-36\%)} \\
    \hline
    \multirow{2}[1]{*}{Perf.} & Vanilla & 82.59 (+0.0\%) & 74.5 (+0.0\%) & 20.5 (+0.0\%) & 83.60 (+0.0\%) & 77.2 (+0.0\%) & 25.0 (+0.0\%) & 79.86 (+0.0\%) & 72.9 (+0.0\%) & 17.9 (+0.0\%) \\
          & Lean & \textbf{83.32 (+0.9\%)} & \textbf{75.0 (+0.7\%)} & \textbf{21.9 (+6.8\%)} & \textbf{84.45 (+1.0\%)} & \textbf{80.1 (+3.8\%)} & \textbf{27.0 (+8.0\%)} & \textbf{81.61 (+2.2\%)} & \textbf{73.2 (+0.4\%)} & \textbf{18.4 (+2.8\%)} \\
    \hline
    \end{tabular}%
  \label{tab:performance_architectures}
\end{table*}%

Our modular blocks can thus be applied to a wide range of state-of-the-art networks and improve significantly their memory consumption while having a positive impact on performance.

\subsection{Ablation study}
\label{subsec:ablation_study}
In this section we report our extensive experiments to assess the importance of each block of our network architectures. Our lean structure allows us to adjust the network architectures by increasing its complexity, either by (i)~adding extra connections or by (ii)~increasing the depth. We analyze our networks along four axes: (i)~the performance (IoU or accuracy) (Table~\ref{tab:accuracyDatasets}), (ii)~the memory footprint, (iii)~the inference time and (iv)~the backward time. Our main experimental findings regarding network efficiency are reported in Table~\ref{tab:efficiency_table} and ablate the impact of our proposed design choices for point processing networks. 

\begin{table*}[t!]
  \footnotesize
  \centering
  \caption{Efficiency of our network architectures measured with a batch size of 8 samples on a Nvidia GTX 2080Ti GPU. All of our lean architectures allow to save a substantial amount of memory on GPU wrt. the PointNet++ baseline from 58\% with \mres to a 67\% decrease with LPN. This latter convolution-type architecture wins on all counts, decreasing both inference time (-41\%) and the length of backward pass (-68\%) by a large spread. Starting from this architecture, the marginal cost of going deep is extremely low: doubling the number of layers in the encoding part of the network increases inference time by 6.3\% on average and the memory consumption by only 3.6\% at most compared to LPN. Please refer to Supplemental for absolute values.}
  \setlength{\tabcolsep}{2.8pt}
    \begin{tabular}{r|ccc|ccc|ccc|ccc}
\cline{2-13}    \multicolumn{1}{r}{} & \multicolumn{3}{c|}{Parameters (M)} & \multicolumn{3}{c|}{Memory Footprint (Gb)} & \multicolumn{3}{c|}{Inference Time (ms)} & \multicolumn{3}{c}{Length Backward pass (ms)} \\
\cline{2-13}  \multicolumn{1}{r}{} & ShapeNet-Part & ScanNet & PartNet & ShapeNet-Part & ScanNet & PartNet & ShapeNet-Part & ScanNet & PartNet & ShapeNet-Part & ScanNet & PartNet \\
    \hline
    PointNet++ & +0.0\% & +0.0\% & +0.0\% & +0.0\% & +0.0\% & +0.0\% & +0.0\% & +0.0\% & +0.0\% & +0.0\% & +0.0\% & +0.0\%     \\
    \hline
    mRes  & -17.0\% & -17.6\% & -16.6\% & -69.3\% & -56.5\% & -47.6\% & +14.8\% & +59.2\% & -19.4\% & -68.8\% & \textbf{-53.8\%} & -63.2\%     \\
    \hline
    mResX & -10.6\% & -10.7\% & -10.1\% & -65.0\% & -53.2\% & -46.3\% & +28.2\% & +60.9\% & -12.5\% & -29.5\% & +0.0\% & -25.4\%     \\
    \hline
    LPN & +13.8\% & +13.4\% & +12.6\% & -75.7\% & \textbf{-66.6\%} & \textbf{-57.9\%} & \textbf{-45.6\%} & \textbf{-30.3\%} & \textbf{-47.9\%} & \textbf{-82.7\%} & -42.3\% & \textbf{-78.9\%}     \\
    \hline
    Deep LPN & \textbf{+54.3\%} & \textbf{+54.0\%} & \textbf{+50.8\%}  & \textbf{-79.1\%} & -65.4\% & -57.0\% & -40.4\% & -25.6\% & -46.5\% & -78.6\% & -11.5\% & -72.4\%    \\\hline
    \end{tabular}
  \label{tab:efficiency_table}
\end{table*}

    \textbf{Memory-efficient Convolutions:} As described in \refsec{conv}, our leanest architeture is equivalent to constraining each PointNet unit to be composed of a single layer network, and turning its operation into a memory-efficient block by removing intermediate activations from memory. In order to get a network of similar size, multiple such units are stacked to reach the same number of layers as the original network. Our convolution-type network wins on all counts, both on performance and efficiency. Indeed, the IoU is increased by 3.4\% on ScanNet and 7.4\% on PartNet compared to \pnpp baseline. Regarding its efficiency, the memory footprint is decreased by 67\% on average while decreasing both inference time (-41\%) and the length of the backward pass (-68\%). These improvements in speed can be seen as the consequence of processing most computations on flattened tensors and thus reducing drastically the complexity compared to \pnpp baseline.

    \textbf{Multi-Resolution:} Processing different resolutions at the same stage of a network has been shown to perform well in shallow networks. Indeed, mixing information at different resolutions helps to capture complex features early in the network. We adopt that approach to design our \mres architecture. Switching from a \pnpp architecture to a multi-resolution setting increases the IoU by 1.0\% on ShapeNet-Part and 5.7\% on PartNet. More crucially, this increase in performance come with more efficiency. Although the inference time is longer (18\% longer on average) due to the extra downsampling and upsampling operations, the architecture is much leaner and reduces memory footprint by 58\%. Training is quicker though due to a 62\% faster backward pass.
    
    \textbf{Cross-links:} Information streams at different resolutions are processed separately and can be seen as complementary. To leverage this synergy, the network is provided with additional links connecting neighborhood resolutions. We experiment on the impact of those cross-resolution links to check their effect on the optimization. At the price of a small impact on memory efficiency (+8\% wrt. \mres) and speed (+7\% on inference time wrt. \mres), the performance can be improved on PartNet, the most complex dataset, with these extra-links by 0.8\%.
    
\section{Conclusion}
\label{sec:conclusion}
We have introduced new generic building blocks for point processing networks, that exhibit  favorable memory, computation, and optimization properties when compared to the current counterparts of state-of-the-art point processing networks. Based on \pnpp, our lean architecture LPN wins on all counts, memory efficiency (-67\% wrt. \pnpp) and speed (-41\% and -68\% on inference time and length of backward pass). Its deep counterpart has a  marginal cost in terms of efficiency and achieves the best IoU on PartNet (+9.7\% over \pnpp). 
Those generic blocks exhibit similar performance on all of the additionally tested architectures producing significantly leaner networks (up to -69\%) and increase in IoU (up to +8.0\%).  Based on our experiments, we anticipate that adding these components to the armament of the deep geometry processing community will allow researchers to train the next generation of point processing networks by leveraging upon the advent of larger shape datasets \cite{DBLP:journals/corr/abs-1812-02713,DBLP:journals/corr/abs-1812-06216}.


 
\newpage
{\small
\bibliographystyle{style/ieee_fullname}

\begin{thebibliography}{10}\itemsep=-1pt

\bibitem{armeni_cvpr16}
Iro Armeni, Ozan Sener, Amir~R. Zamir, Helen Jiang, Ioannis Brilakis, Martin
  Fischer, and Silvio Savarese.
\newblock 3d semantic parsing of large-scale indoor spaces.
\newblock In {\em Proceedings of the IEEE International Conference on Computer
  Vision and Pattern Recognition}, 2016.

\bibitem{Atzmon:2018:PCN:3197517.3201301}
Matan Atzmon, Haggai Maron, and Yaron Lipman.
\newblock Point convolutional neural networks by extension operators.
\newblock {\em ACM Trans. Graph.}, 37(4):71:1--71:12, July 2018.

\bibitem{matterport17}
Angel Chang, Angela Dai, Thomas Funkhouser, Maciej Halber, Matthias
  Nie√üner, Manolis Savva, Shuran Song, Andy Zeng, and Yinda Zhang.
\newblock Matterport3d: Learning from rgb-d data in indoor environments.
\newblock 09 2017.

\bibitem{shapenet2015}
Angel~X. Chang, Thomas Funkhouser, Leonidas Guibas, Pat Hanrahan, Qixing Huang,
  Zimo Li, Silvio Savarese, Manolis Savva, Shuran Song, Hao Su, Jianxiong Xiao,
  Li Yi, and Fisher Yu.
\newblock {ShapeNet: An Information-Rich 3D Model Repository}.
\newblock Technical Report arXiv:1512.03012 [cs.GR], Stanford University ---
  Princeton University --- Toyota Technological Institute at Chicago, 2015.

\bibitem{ChenXZG16}
Tianqi Chen, Bing Xu, Chiyuan Zhang, and Carlos Guestrin.
\newblock Training deep nets with sublinear memory cost.
\newblock {\em CoRR}, abs/1604.06174, 2016.

\bibitem{dai2017scannet}
Angela Dai, Angel~X. Chang, Manolis Savva, Maciej Halber, Thomas Funkhouser,
  and Matthias Nie{\ss}ner.
\newblock Scannet: Richly-annotated 3d reconstructions of indoor scenes.
\newblock In {\em Proc. Computer Vision and Pattern Recognition (CVPR), IEEE},
  2017.

\bibitem{gadelha2018multiresolution}
Matheus Gadelha, Rui Wang, and Subhransu Maji.
\newblock Multiresolution tree networks for 3d point cloud processing.
\newblock In {\em Proceedings of the European Conference on Computer Vision
  (ECCV)}, pages 103--118, 2018.

\bibitem{GomezRUG17}
Aidan~N. Gomez, Mengye Ren, Raquel Urtasun, and Roger~B. Grosse.
\newblock The reversible residual network: Backpropagation without storing
  activations.
\newblock {\em CoRR}, abs/1707.04585, 2017.

\bibitem{NIPS2016_6221}
Audrunas Gruslys, Remi Munos, Ivo Danihelka, Marc Lanctot, and Alex Graves.
\newblock Memory-efficient backpropagation through time.
\newblock In D.~D. Lee, M. Sugiyama, U.~V. Luxburg, I. Guyon, and R. Garnett,
  editors, {\em Advances in Neural Information Processing Systems 29}, pages
  4125--4133. Curran Associates, Inc., 2016.

\bibitem{GuerreroEtAl:PCPNet:EG:2018}
Paul Guerrero, Yanir Kleiman, Maks Ovsjanikov, and Niloy~J. Mitra.
\newblock {PCPNet}: Learning local shape properties from raw point clouds.
\newblock {\em CGF}, 37(2):75--85, 2018.

\bibitem{DBLP:journals/corr/HariharanAGM14a}
Bharath Hariharan, Pablo~Andr{\'{e}}s Arbel{\'{a}}ez, Ross~B. Girshick, and
  Jitendra Malik.
\newblock Hypercolumns for object segmentation and fine-grained localization.
\newblock {\em CoRR}, abs/1411.5752, 2014.

\bibitem{HeZRS15}
Kaiming He, Xiangyu Zhang, Shaoqing Ren, and Jian Sun.
\newblock Deep residual learning for image recognition.
\newblock 2016.

\bibitem{hermosilla2018mccnn}
P. Hermosilla, T. Ritschel, P-P Vazquez, A. Vinacua, and T. Ropinski.
\newblock Monte carlo convolution for learning on non-uniformly sampled point
  clouds.
\newblock {\em ACM Transactions on Graphics (Proceedings of SIGGRAPH Asia
  2018)}, 37(6), 2018.

\bibitem{densenet}
Gao Huang, Zhuang Liu, and Kilian~Q. Weinberger.
\newblock Densely connected convolutional networks.
\newblock {\em CoRR}, abs/1608.06993, 2016.

\bibitem{Jia14}
Yangqing Jia.
\newblock {\em Learning Semantic Image Representations at a Large Scale}.
\newblock PhD thesis, University of California, Berkeley, {USA}, 2014.

\bibitem{DBLP:journals/corr/KeMY16}
Tsung{-}Wei Ke, Michael Maire, and Stella~X. Yu.
\newblock Neural multigrid.
\newblock {\em CoRR}, abs/1611.07661, 2016.

\bibitem{DBLP:journals/corr/abs-1812-06216}
Sebastian Koch, Albert Matveev, Zhongshi Jiang, Francis Williams, Alexey
  Artemov, Evgeny Burnaev, Marc Alexa, Denis Zorin, and Daniele Panozzo.
\newblock {ABC:} {A} big {CAD} model dataset for geometric deep learning.
\newblock {\em CoRR}, abs/1812.06216, 2018.

\bibitem{KrizhevskyNIPS2013}
A. Krizhevsky, I. Sutskever, and G.~E. Hinton.
\newblock Imagenet classification with deep convolutional neural networks.
\newblock In {\em NIPS}, 2013.

\bibitem{li2019deepgcns}
Guohao Li, Matthias Müller, Ali Thabet, and Bernard Ghanem.
\newblock Deepgcns: Can gcns go as deep as cnns?, 2019.

\bibitem{li2018sonet}
Jiaxin Li, Ben~M Chen, and Gim Hee~Lee.
\newblock So-net: Self-organizing network for point cloud analysis.
\newblock pages 9397--9406, 2018.

\bibitem{li2018pointcnn}
Yangyan Li, Rui Bu, Mingchao Sun, Wei Wu, Xinhan Di, and Baoquan Chen.
\newblock Pointcnn: Convolution on x-transformed points.
\newblock 2018.

\bibitem{DBLP:journals/corr/abs-1904-07601}
Yongcheng Liu, Bin Fan, Shiming Xiang, and Chunhong Pan.
\newblock Relation-shape convolutional neural network for point cloud analysis.
\newblock {\em CoRR}, 2019.

\bibitem{DBLP:journals/corr/abs-1812-02713}
Kaichun Mo, Shilin Zhu, Angel~X. Chang, Li Yi, Subarna Tripathi, Leonidas~J.
  Guibas, and Hao Su.
\newblock Partnet: {A} large-scale benchmark for fine-grained and hierarchical
  part-level 3d object understanding.
\newblock {\em CoRR}, abs/1812.02713, 2018.

\bibitem{QiEtAl:Pointnet:CVPR:2017}
Charles~R Qi, Hao Su, Kaichun Mo, and Leonidas~J Guibas.
\newblock Pointnet: Deep learning on point sets for 3d classification and
  segmentation.
\newblock {\em CVPR}, 1(2):4, 2017.

\bibitem{QiEtAl:Pointnet++:NIPS:2017}
Charles~Ruizhongtai Qi, Li Yi, Hao Su, and Leonidas~J Guibas.
\newblock Pointnet++: Deep hierarchical feature learning on point sets in a
  metric space.
\newblock In {\em NIPS}, pages 5099--5108, 2017.

\bibitem{simonyan2014very}
Karen Simonyan and Andrew Zisserman.
\newblock Very deep convolutional networks for large-scale image recognition.
\newblock 2015.

\bibitem{su2018splatnet}
Hang Su, Varun Jampani, Deqing Sun, Subhransu Maji, Evangelos Kalogerakis,
  Ming-Hsuan Yang, and Jan Kautz.
\newblock Splatnet: Sparse lattice networks for point cloud processing.
\newblock pages 2530--2539, 2018.

\bibitem{su15mvcnn}
Hang Su, Subhransu Maji, Evangelos Kalogerakis, and Erik~G. Learned{-}Miller.
\newblock Multi-view convolutional neural networks for 3d shape recognition.
\newblock In {\em Proc. ICCV}, 2015.

\bibitem{SzegedyLJSRAEVR14}
Christian Szegedy, Wei Liu, Yangqing Jia, Pierre Sermanet, Scott~E. Reed,
  Dragomir Anguelov, Dumitru Erhan, Vincent Vanhoucke, and Andrew Rabinovich.
\newblock Going deeper with convolutions.
\newblock {\em CoRR}, abs/1409.4842, 2014.

\bibitem{DBLP:journals/corr/abs-1904-08889}
Hugues Thomas, Charles~R. Qi, Jean{-}Emmanuel Deschaud, Beatriz Marcotegui,
  Fran{\c{c}}ois Goulette, and Leonidas~J. Guibas.
\newblock Kpconv: Flexible and deformable convolution for point clouds.
\newblock {\em CoRR}, 2019.

\bibitem{Wang-2017-OCNN}
Peng-Shuai Wang, Yang Liu, Yu-Xiao Guo, Chun-Yu Sun, and Xin Tong.
\newblock {O-CNN: Octree-based Convolutional Neural Networks for 3D Shape
  Analysis}.
\newblock {\em ACM Transactions on Graphics (SIGGRAPH)}, 36(4), 2017.

\bibitem{Wang-2018-AOCNN}
Peng-Shuai Wang, Chun-Yu Sun, Yang Liu, and Xin Tong.
\newblock {Adaptive O-CNN: A Patch-based Deep Representation of 3D Shapes}.
\newblock {\em ACM Transactions on Graphics (SIGGRAPH Asia)}, 37(6), 2018.

\bibitem{WangEtAl:DGCNN:arxiv:2018}
Yue Wang, Yongbin Sun, Ziwei Liu, Sanjay~E Sarma, Michael~M Bronstein, and
  Justin~M Solomon.
\newblock Dynamic graph cnn for learning on point clouds.
\newblock {\em arXiv preprint arXiv:1801.07829}, 2018.

\bibitem{DBLP:journals/corr/XieGDTH16}
Saining Xie, Ross~B. Girshick, Piotr Doll{\'{a}}r, Zhuowen Tu, and Kaiming He.
\newblock Aggregated residual transformations for deep neural networks.
\newblock {\em CoRR}, abs/1611.05431, 2016.

\bibitem{xu2018spidercnn}
Yifan Xu, Tianqi Fan, Mingye Xu, Long Zeng, and Yu Qiao.
\newblock Spidercnn: Deep learning on point sets with parameterized
  convolutional filters.
\newblock In {\em Proceedings of the European Conference on Computer Vision
  (ECCV)}, pages 87--102, 2018.

\bibitem{Yin:2018:PBP:3197517.3201288}
Kangxue Yin, Hui Huang, Daniel Cohen-Or, and Hao Zhang.
\newblock P2p-net: Bidirectional point displacement net for shape transform.
\newblock {\em ACM TOG}, 37(4):152:1--152:13, July 2018.

\bibitem{yu2018ecnet}
Lequan Yu, Xianzhi Li, Chi-Wing Fu, Daniel Cohen-Or, and Pheng-Ann Heng.
\newblock Ec-net: an edge-aware point set consolidation network.
\newblock pages 386--402, 2018.

\bibitem{yu2018pu}
Lequan Yu, Xianzhi Li, Chi-Wing Fu, Daniel Cohen-Or, and Pheng-Ann Heng.
\newblock Pu-net: Point cloud upsampling network.
\newblock In {\em CVPR}, 2018.

\bibitem{wrnet}
Sergey Zagoruyko and Nikos Komodakis.
\newblock Wide residual networks.
\newblock {\em CoRR}, abs/1605.07146, 2016.

\bibitem{zaheer2017deepsets}
Manzil Zaheer, Satwik Kottur, Siamak Ravanbakhsh, Barnabas Poczos, Ruslan~R
  Salakhutdinov, and Alexander~J Smola.
\newblock Deep sets.
\newblock 2017.

\bibitem{ZeilerF14}
Matthew~D. Zeiler and Rob Fergus.
\newblock Visualizing and understanding convolutional networks.
\newblock 2014.

\bibitem{zhang-shellnet-iccv19}
Zhiyuan Zhang, Binh-Son Hua, and Sai-Kit Yeung.
\newblock Shellnet: Efficient point cloud convolutional neural networks using
  concentric shells statistics.
\newblock In {\em International Conference on Computer Vision (ICCV)}, 2019.

\end{thebibliography}

}

\clearpage

\beginsupplement

\section{Details on evaluation results}

\subsection{Datasets}
We evaluate our networks on the point cloud segmentation task on three different datasets, ordered by increasing complexity:
\begin{itemize}
    \item ShapeNet-Part~\cite{shapenet2015}: CAD models of $16$ different object categories composed of $50$ labeled parts. The dataset provides $13,998$ samples for training and $2,874$ samples for evaluation. Point segmentation performance is assessed using the mean point Intersection over Union (mIoU).
    \item ScanNet~\cite{dai2017scannet}: Scans of real 3D scenes (scanned and reconstructed indoor scenes) composed of $21$ semantic parts. The dataset provides $1,201$ samples for training and $312$ samples for evaluation. We follow the same protocol as in \cite{QiEtAl:Pointnet:CVPR:2017} and report both the voxel accuracy and the part Intersection over Union (pIoU).
    \item PartNet \cite{DBLP:journals/corr/abs-1812-02713}: Large collection of CAD models of $17$ object categories composed of $251$ labeled parts. The dataset provides $17,119$ samples for training, $2,492$ for validation and $4,895$ for evaluation. The dataset provides a benchmark for three different tasks: fine-grained semantic segmentation, hierarchical semantic segmentation and instance segmentation. We report on the first task to evaluate our networks on a more challenging segmentation task using the same part Intersection over Union~(pIoU) as in ScanNet.
\end{itemize}

\begin{table}[b!]
  \centering
  \caption{Summary of the impact of our module implants in five different networks on ShapeNet-Part. The impact is measured by four metrics: (i) memory footprint, (ii) IoU, (iii) inference time and (iv) backward time. With all tested architectures, our lean modules decrease the memory footprint while allowing small improvements in terms of IoU. The impact on inference time depends on the choice of the network but can range from positive impact to a small slowdown.}
    \begin{tabular}{r|c|c|c|c}
\cline{2-5}    \multicolumn{1}{r}{} & Memory & IoU & Inference & Backward \\
    \hline
    PN++  & \textbf{-76\%} & \textbf{+1.2\%} & \textbf{-46\%} & -\textbf{83\%} \\
    \hline
    DGCNN & \textbf{-69\%} & \textbf{+0.9\%} & \textbf{-22\%} & \textbf{-49\%} \\
    \hline
    SCNN  & \textbf{-28\%} & \textbf{+2.2\%} & +27\% & +193\% \\
    \hline
    PointCNN & \textbf{-56\%} & \textbf{+1.0\%} & +35\%  & +71\% \\
    \hline
    \end{tabular}%
  \label{tab:summary_table}%
\end{table}%

\subsection{Evaluation metrics}
\label{appendix:evaluation_metrics}
To report our segmentation results, we use two versions of the Intersection over Union metric:
\begin{itemize}
    \item mIoU: To get the per sample mean-IoU, the IoU is first computed for each part belonging to the given object category, whether or not the part is in the sample. Then, those values are averaged across the parts. If a part is neither predicted nor in the ground truth, the IoU of the part is set to 1 to avoid this indefinite form. The mIoU obtained for each sample is then averaged to get the final score as,
    \begin{equation*}
    \text{mIoU}=\frac{1}{n_{\text{samples}}}\sum_{s\in\text{samples}}\frac{1}{n_{\text{parts}}^{\text{cat(s)}}}\sum_{p^i\in\mathcal{P}_{\text{cat(s)}}} \text{IoU}_s(p^i)
    \end{equation*}
    with $n_{\text{samples}}$ the number of samples in the dataset, $\text{cat(s)}$, $n_{\text{parts}}^{\text{cat(s)}}$ and $\mathcal{P}_{\text{cat(s)}}$ the object category where $s$ belongs, the number of parts in this category and the sets of its parts respectively. $\text{IoU}_s(p^i)$ is the IoU of part $p^i$ in sample $s$.
    
    \item pIoU: The part-IoU is computed differently. The IoU per part is first computed over the whole dataset and then, the values obtained are averaged across the parts as,
    \begin{equation*}
    \text{pIoU} = \frac{1}{n_{\text{parts}}}\sum_{p\in\text{parts}}\frac{\sum_{s\in\text{samples}}\text{I}_s(p^i)}{\sum_{s\in\text{samples}}\text{U}_s(p^i)}
    \end{equation*}
    with $n_{\text{parts}}$ the number of parts in the dataset, $\text{I}_s(p^i)$ and $\text{U}_s(p^i)$ the intersection and union for samples $s$ on part $p^i$ respectively.
\end{itemize}
To take into account the randomness of point cloud sampling when performing coarsening, we use the average of `N' forward passes to decide on the final segmentation during evaluation when relevant.

\begin{table*}[ht!]
  \small
  \centering
  \setlength{\tabcolsep}{2.8pt}
  \caption{Per-category performance mIoU on ShapeNet-Part (Top) and pIoU on PartNet (Bottom) based on a training on each whole dataset all at once. On ShapeNet-Part, all of our network architectures outperform \pnpp baseline by at least +1.0\%. Our deep architecture still improves the performance of its shallower counterpart by a small margin of +0.1\%. On PartNet, the fine details of the segmentation and the high number of points to process make the training much more complex than previous datasets. \pnpp, here, fails to capture enough features to segment the objects properly. Our different architectures outperform \pnpp with a spread of at least +2.0\% (+5.7\% increase). With this more complex dataset, deeper networks become significantly better: our Deep LPN network achieves to increase pIoU by +9.7\% over \pnpp baseline, outperforming its shallow counterpart by +2.1\%.}
  
\begin{tabular}{r|ccccccccccccccccc}
\cline{2-18}    \multicolumn{1}{c}{} & Tot./Av. & Aero  & Bag   & Cap   & Car   & Chair & Ear   & Guitar & Knife & Lamp  & Laptop & Motor & Mug   & Pistol & Rocket & Skate & Table \\
\hline
No. Samples & 13998 & 2349  & 62    & 44    & 740   & 3053  & 55    & 628   & 312   & 1261  & 367   & 151   & 146   & 234   & 54    & 121   & 4421 \\
\hline
    PN++ & 84.60 & 82.7 & 76.8 & 84.4 & 78.7 & 90.5 & 72.3 & 90.5 & 86.3 & 82.9 & \textbf{96.0} & 72.4 & 94.3 & 80.5 & \textbf{62.8} & 76.3 & 81.2 \\
\hline
    mRes & 85.47 & \textbf{83.7} & 77.1 & 85.4 & 79.6 & \textbf{91.2} & 73.4 & 91.6 & \textbf{88.1} & 84.1 & 95.6 & 75.1 & 95.1 & 81.4 & 59.7 & 76.9 & 82.1 \\
\hline
    mResX & 85.42 & 83.1 & 77.0 & 84.8 & 79.7 & 91.0 & 67.8 & 91.5 & 88.0 & 84.1 & 95.7 & 74.6 & \textbf{95.4} & 82.4 & 57.1 & 77.0 & 82.3 \\
\hline
    LPN & 85.65 & 83.3 & 77.2 & \textbf{87.8} & 80.6 & 91.1 & 72.0 & \textbf{91.8} & \textbf{88.1} & 84.6 & 95.8 & \textbf{75.8} & 95.1 & \textbf{83.6} & 60.7 & 75.0 & 82.4 \\
\hline
    Deep LPN & \textbf{85.66} & 82.8 & \textbf{79.2} & 82.7 & \textbf{80.9} & 91.1 & \textbf{75.4} & 91.6 & \textbf{88.1} & \textbf{84.9} & 95.3 & 73.1 & 95.1 & 83.3 & 61.6 & \textbf{77.7} & \textbf{82.6} \\
\hline
    \end{tabular}
    \newline
    \vspace*{5pt}
    \newline
    \begin{tabular}{r|cccccccccccccccccc}
\cline{2-19}   \multicolumn{1}{r}{} & \multicolumn{1}{c}{Tot./Av.} & \multicolumn{1}{c}{Bed} & \multicolumn{1}{c}{Bott} & \multicolumn{1}{c}{Chair} & \multicolumn{1}{c}{Clock} & \multicolumn{1}{c}{Dish} & \multicolumn{1}{c}{Disp} & \multicolumn{1}{c}{Door} & \multicolumn{1}{c}{Ear} & \multicolumn{1}{c}{Fauc} & \multicolumn{1}{c}{Knife} & \multicolumn{1}{c}{Lamp} & \multicolumn{1}{c}{Micro} & \multicolumn{1}{c}{Frid} & \multicolumn{1}{c}{Storage} & \multicolumn{1}{c}{Table} & \multicolumn{1}{c}{Trash} & \multicolumn{1}{c}{Vase} \\
    \hline
    No. samples & 17119 & 133   & 315   & 4489  & 406   & 111   & 633   & 149   & 147   & 435   & 221   & 1554  & 133   & 136   & 1588  & 5707  & 221   & 741 \\
    \hline
    PN++ & 35.2 & 30.1 & 32.0 & 39.5 & 30.3 & 29.1 & 81.4 & 31.4 & 35.4 & 46.6 & 37.1 & 25.1 & 31.5 & 32.6 & 40.5 & 34.9 & 33.0 & 56.3 \\
    \hline
    mRes & 37.2 & 29.6 & 32.7 & 40.0 & 34.3 & 29.9 & 80.2 & \textbf{35.0} & \textbf{50.0} & \textbf{56.5} & 41.0 & 26.5 & \textbf{33.9} & \textbf{35.1} & 41.0 & 35.4 & 35.3 & 57.7\\
    \hline
    mResX & 37.5 & 32.0 & 37.9 & 40.4 & 30.2 & 31.8 & 80.9 & 34.0 & 43.0 & 54.3 & 42.6 & 26.8 & 33.1 & 31.8 & \textbf{41.2} & \textbf{36.5} & 40.8 & 57.2 \\
    \hline
    LPN & 37.8 & \textbf{33.2} & 40.7 & 40.8 & \textbf{35.8} & 31.9 & 81.2 & 33.6 & 48.4 & 54.3 & 41.8 & 26.8 & 31.0 & 32.2 & 40.6 & 35.4 & 41.1 & 57.2 \\
    \hline
    Deep LPN & \textbf{38.6} & 29.5 & \textbf{42.1} & \textbf{41.8} & 34.7 & \textbf{33.2} & \textbf{81.6} & 34.8 & 49.6 & 53.0 & \textbf{44.8} & \textbf{28.4} & 33.5 & 32.3 & 41.1 & 36.3 & \textbf{43.1} & \textbf{57.8} \\
    \hline
    \end{tabular}
  \label{tab:perclass_accuracy}
\end{table*}

\subsection{Summary of the impact of our module}
We experiment on four different networks that all exhibit diverse approach to point operation: (i) PointNet++~\cite{QiEtAl:Pointnet++:NIPS:2017}, (ii) Dynamic Graph CNN~\cite{WangEtAl:DGCNN:arxiv:2018}, (iii) SpiderCNN~\cite{xu2018spidercnn}, (iv) PointCNN~\cite{li2018pointcnn} .
As detailed in Table~\ref{tab:summary_table}, our lean blocks, being modular and generic, can not only increase the memory efficiency of that wide range of networks but can as well improve their accuracy. The effect of our blocks on inference time does vary with the type of network, from a positive impact to a small slowdown.

\subsection{Detailed results from the paper}
The following section provides more details on the evaluation experiments introduced in the paper. We present the per-class IoU on both ShapeNet-Part and PartNet datasets in Table~\ref{tab:perclass_accuracy} for each of the \pnpp based architecture. Due to the high number of points per sample and the level of details of the segmentation, PartNet can be seen as much more complex than ShapeNet-Part.

On PartNet, the spread between an architecture with an improved information flow and a vanilla one becomes significant. Our \pnpp based networks perform consistently better than the original architecture on each of the PartNet classes. Increasing the depth of the network allows to achieve a higher accuracy on the most complex classes such as Chairs or Lamps composed of 38 and 40 different part categories respectively. Our deep architecture is also able to better capture the boundaries between parts and thus to predict the right labels very close from part edges. When a sample is itself composed of many parts, having a deep architecture is a significant advantage.

As additional reference, we provide on Table~\ref{tab:perclass_training} the performance of our lean blocks applied to three architectures when training one network per-object category on PartNet, trained on \textit{Chairs} and \textit{Tables} as they represent 60\% of the dataset.

\begin{table}[t!]
\footnotesize
  \centering
  \caption{Per-class IoU on PartNet when training a separate network for each category, evaluated for three different architectures for \textit{Chairs} and \textit{Tables} (60\% of the whole dataset). Our lean networks achieve here similar performance as their vanilla counterpart while delivering significant savings in memory.}
    \begin{tabular}{r|r|c|c}
\cline{3-4}    \multicolumn{1}{r}{} & \multicolumn{1}{r}{} & Chair & Table \\
    \hline
    \multirow{2}[2]{*}{DGCNN} & Vanilla & \textbf{29.2 (+0.0\%)} & 22.5 (+0.0\%) \\
          & Lean & 24.2 (-17.1\%) & \textbf{28.9 (+28.4\%)} \\
    \hline
    \multirow{2}[2]{*}{SCNN} & Vanilla & 30.8 (+0.0\%) & \textbf{21.3 (+0.0\%)} \\
          & Lean & \textbf{31.1 (+1.0\%)} & 21.2 (-0.5\%) \\
    \hline
    \multirow{2}[2]{*}{PointCNN} & Vanilla & 40.4 (+0.0\%) & 32.1 (+0.0\%) \\
          & Lean & \textbf{41.4 (+2.5\%)} & \textbf{33.1 (+3.1\%)} \\
    \hline
    \end{tabular}%

  \label{tab:perclass_training}%
\end{table}%

\begin{table}[t!]
  \tiny
  \scriptsize
  \caption{Memory and speed efficiency of our deep network Deep LPN with respect to two different implementations of DeepGCNs. Our network wins on all counts and successfully reduces the memory (-~75\%) and increases the speed (-~48\% and -~89\% for the inference and the backward time respectively).}
  \setlength{\tabcolsep}{4.2pt}
    \begin{tabular}{r|c|c|c}
\cline{2-4}    \multicolumn{1}{r}{} & Memory (Gb) & \multicolumn{1}{l|}{Inference Time (ms)} & \multicolumn{1}{l}{Backward Time (ms)} \\
    \hline
    DeepGCN (Dense) & 8.56  & 664   & 1088 \\
    DeepGCN (Sparse) & 10.00 & 1520  & 445 \\
    \hline
    Deep LPN & \textbf{2.18 (-75\%)}  & \textbf{345 (-48\%)}   & \textbf{67 (-85\%)} \\
    \hline
    \end{tabular}%
  \label{tab:depgcn_efficiency}%
\end{table}%

\begin{table*}[htbp]
  \centering
  \footnotesize
  \caption{Efficiency of our network architectures measured with a batch size of 8 samples or less on a Nvidia GTX 2080Ti GPU. All of our lean architectures allow to save a substantial amount of memory on GPU wrt. the PointNet++ baseline from 58\% with \mres to a 67\% decrease with LPN. This latter convolution-type architecture wins on all counts, decreasing both inference time (-41\%) and the length of backward pass (-68\%) by a large spread. Starting form this architecture, the marginal cost of going deep is extremely low: doubling the number of layers in the encoding part of the network increases inference time by 6.3\% on average and the memory consumption by only 3.6\% at most compared to LPN). When used in conjunction with other base architectures, similar memory savings are achieved by our blocks with low impact on inference time.}
  \setlength{\tabcolsep}{2.8pt}
      \begin{tabular}{r|ccc|ccc|ccc|ccc}
\cline{2-13}    \multicolumn{1}{r}{} & \multicolumn{3}{c|}{Efficiency (\%)} & \multicolumn{3}{c|}{Memory Footprint (Gb)} & \multicolumn{3}{c|}{Inference Time (ms)} & \multicolumn{3}{c}{Length Backward pass (ms)} \\
\cline{2-13}  \multicolumn{1}{r}{} & ShapeNet-Part & ScanNet & PartNet & ShapeNet-Part & ScanNet & PartNet & ShapeNet-Part & ScanNet & PartNet & ShapeNet-Part & ScanNet & PartNet \\
    \hline
    PointNet++ & 84.60 & 80.5 & 35.2 & 6.80 & 6.73 & 7.69 & 344 & 238 & 666 & 173 & 26 & 185     \\
    \hline
    mRes  & 85.47 & 79.4 & 37.2 & 2.09 & 2.93 & 4.03 & 395 & 379 & 537 & 54 & \textbf{12} & 68     \\
    \hline
    mResX & 85.42 & 79.5 & 37.5 & 2.38 & 3.15 & 4.13 & 441 & 383 & 583 & 122 & 26 & 138     \\
    \hline
    LPN & 85.65 & \textbf{83.2} & 37.8 & 1.65 & \textbf{2.25} & \textbf{3.24} & \textbf{187} & \textbf{166} & \textbf{347} & \textbf{30} & 15 & \textbf{39}     \\
    \hline
    Deep LPN & \textbf{85.66} & 82.2 & \textbf{38.6} & \textbf{1.42} & 2.33 & 3.31 & 205 & 177 & 356 & 37 & 23 & 51    \\\hline
    \end{tabular}
    \medbreak
    \setlength{\tabcolsep}{2.0pt}
    \begin{tabular}{r|r|ccc|ccc|ccc|ccc}
\cline{3-14}    \multicolumn{1}{r}{} & \multicolumn{1}{r}{} & \multicolumn{3}{c|}{Efficiency (\%)} & \multicolumn{3}{c|}{Memory Footprint (Gb)} & \multicolumn{3}{c|}{Inference Time (ms)} & \multicolumn{3}{c}{Length Backward pass (ms)} \\
\cline{3-14}    \multicolumn{1}{r}{} & \multicolumn{1}{r}{} & ShapeNet-Part & ScanNet & PartNet & ShapeNet-Part & ScanNet & PartNet & ShapeNet-Part & ScanNet & PartNet & ShapeNet-Part & ScanNet & PartNet \\
    \hline
    \multirow{2}[0]{*}{DGCNN} & Vanilla & 82.59 & 74.5 & 20.5 & 2.62  & 7.03  & 9.50  & 41    & 194   & 216   & 41    & 82    & 104 \\
          & Lean & \textbf{83.32} & \textbf{75.0} & \textbf{21.9} & \textbf{0.81}  & \textbf{3.99}  & \textbf{5.77}  & \textbf{32}    & \textbf{158}   & \textbf{168}   & \textbf{21}    & \textbf{45}    & \textbf{57} \\
    \hline
    \multirow{2}[0]{*}{SCNN} & Vanilla & 79.86 & 72.9 & 17.9 & 1.09  & 4.33  & 5.21  & \textbf{22}    & \textbf{279}   & \textbf{142}   & \textbf{45}    & \textbf{99}    & \textbf{249} \\
          & Lean & \textbf{81.61} & \textbf{73.2} & \textbf{18.4} & \textbf{0.79}  & \textbf{3.25}  & \textbf{3.33}  & 28    & 281    & 150   & 132   & 443   & 637 \\
    \hline
    \multirow{2}[0]{*}{PointCNN} & Vanilla & 83.60 & 77.2 & 25.0 & 4.54  & 5.18  & 6.83  & \textbf{189}   & \textbf{229}   & \textbf{228}   & \textbf{109}   & \textbf{71}    & \textbf{77} \\
      & Lean & \textbf{84.45} & \textbf{80.1} & \textbf{27.0} & \textbf{1.98}  & \textbf{3.93}  & \textbf{5.55}  & 256   & 278   & 263   & 186   & 225   & 208 \\
    \hline
    \end{tabular}%

  \label{tab:efficiency_table_absolutevalues}
\end{table*}

For reference, we provide as well the absolute values for the efficiency of the previous networks measured by three different metrics on Table~\ref{tab:depgcn_efficiency} and Table~\ref{tab:efficiency_table_absolutevalues}: (i) memory footprint, (ii) inference time and (iii) length of backward pass. Our lean architectures consistently reduce the memory consumption of their vanilla couterparts while having a very low impact on inference time. When compared to DeepGCNs, our Deep LPN architecture wins on all counts by achieving the same performance while requiring less memory (-75\%) and shorter inference (-48\%) and backward (-89\%) time.

\section{Design of our architectures}
\begin{figure*}[ht!]
    \centering
    \includegraphics[width=\textwidth]{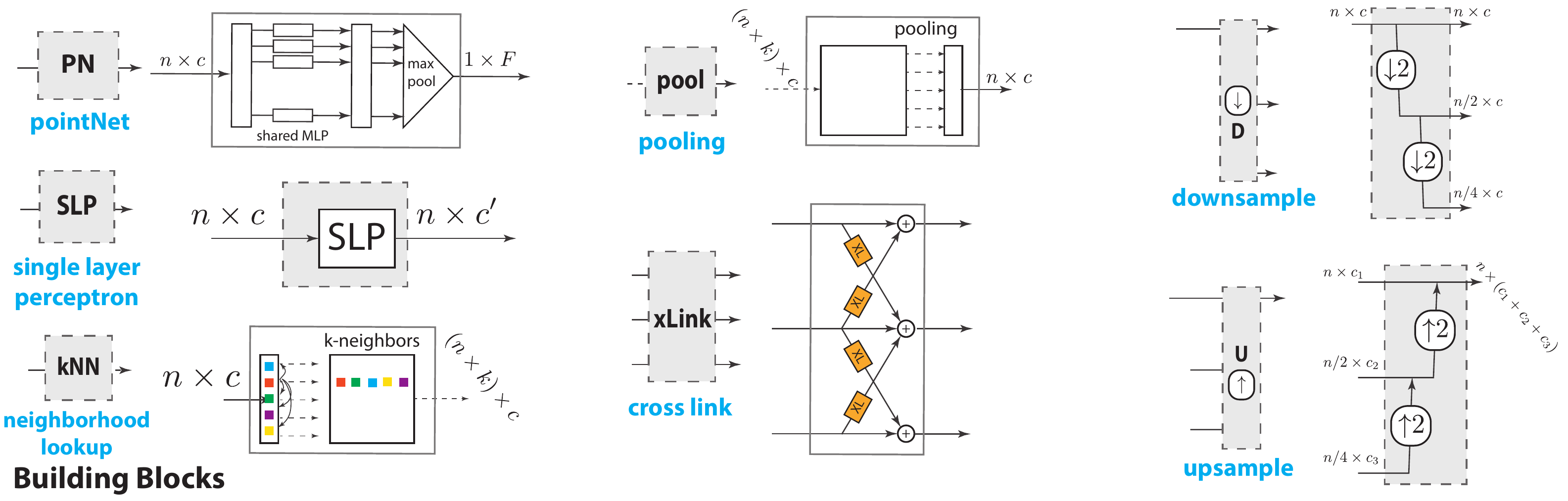}
    \caption{
     Elementary building blocks for point processing. Apart from standard neighborhood lookup, pooling and SLP layers, we introduce cross-link layers across scales, and propose multi-resolution up/down sampling blocks for point processing. PointNet module combines a stack of shared SLP (forming an MLP) to lift individual points \iffalse to higher dimensional features \fi and then performs permutation-invariant local pooling.}
    \label{fig:pipeline_blocks}
\end{figure*}
In this section, we provide more details about how we design our lean architectures to ensure reproducible results for the following architectures, (i) PointNet++~\cite{QiEtAl:Pointnet++:NIPS:2017}, (ii) Dynamic Graph CNN~\cite{WangEtAl:DGCNN:arxiv:2018}, (iii) SpiderCNN~\cite{xu2018spidercnn}, (iv) PointCNN~\cite{li2018pointcnn} . We implement each networks in Pytorch following the original code in Tensorflow and we implant our blocks directly within those networks.

\subsection{PointNet++ based architectures}
\label{appendix:pointnet_based_details}
To keep things simple and concise in this section, we adopt the following notations:
\begin{itemize}
    \item S(n): Sampling layer of n points;
    \item rNN(r): query-ball of radius r;
    \item MaxP: Max Pooling along the neighborhood axis;
    \item $\bigoplus$: Multi-resolution combination;
    \item Lin(s): Linear unit of s neurons;
    \item Drop(p): Dropout layer with a probability p to zero a neuron.
\end{itemize}
Inside our architectures, every downsampling module is itself based on FPS to decrease the resolution of the input point cloud. To get back to the original resolution, upsampling layers proceed to linear interpolation (Interp) in the spatial space using the $K_u=3$ closest neighbors. To generate multiple resolutions of the same input point cloud, a downsampling ratio of 2 is used for every additional resolution. 

\subsubsection{\pnpp}
In all our experiments, we choose to report the performance of the multi-scale \pnpp (MSG PN++) as it is reported to beat its alternative versions in the original paper on all tasks. We code our own implementation of \pnpp in Pytorch and choose the same parameters as in the original code.
\medbreak
For segmentation task, the architecture is designed as follow:

\noindent\underline{Encoding1:}\\
S($512$)$\rightarrow\begin{bmatrix}\text{rNN}(.1)\rightarrow\text{mLP}([32,32,64])\rightarrow \text{MaxP}\\\text{rNN}(.2)\rightarrow\text{mLP}([64,64,128])\rightarrow \text{MaxP} \\\text{rNN}(.4)\rightarrow\text{mLP}([64,96,128])\rightarrow \text{MaxP}\end{bmatrix}\bigoplus$\\
\underline{Encoding2:}\\
S($128$)$\rightarrow\begin{bmatrix}\text{rNN}(.2)\rightarrow\text{mLP}([64,64,128])\rightarrow \text{MaxP} \\\text{rNN}(.4)\rightarrow\text{mLP}([128,128,256])\rightarrow \text{MaxP} \\\text{rNN}(.8)\rightarrow\text{mLP}([128,128,256])\rightarrow \text{MaxP}\end{bmatrix}\bigoplus$\\
\underline{Encoding3:}\\
S($1$)$\rightarrow\text{mLP}([256,512,1024])\rightarrow$ MaxP\\
\underline{Decoding1:} Interp($3$)$\rightarrow\text{mLP}([256,256])$\\
\underline{Decoding2:} Interp($3$)$\rightarrow\text{mLP}([256,128])$\\
\underline{Decoding3:} Interp($3$)$\rightarrow\text{mLP}([128,128])$\\
\underline{Classification:} Lin($512$)$\rightarrow$ Drop($.7$)$\rightarrow$ Lin($\text{nb}_{\text{classes}}$)\\
We omit here skiplinks for sake of clarity: they connect encoding and decoding modules at the same scale level.

\subsubsection{\mres}
\label{appendix:mres_details}
\begin{figure}
    \includegraphics[width=1\columnwidth]{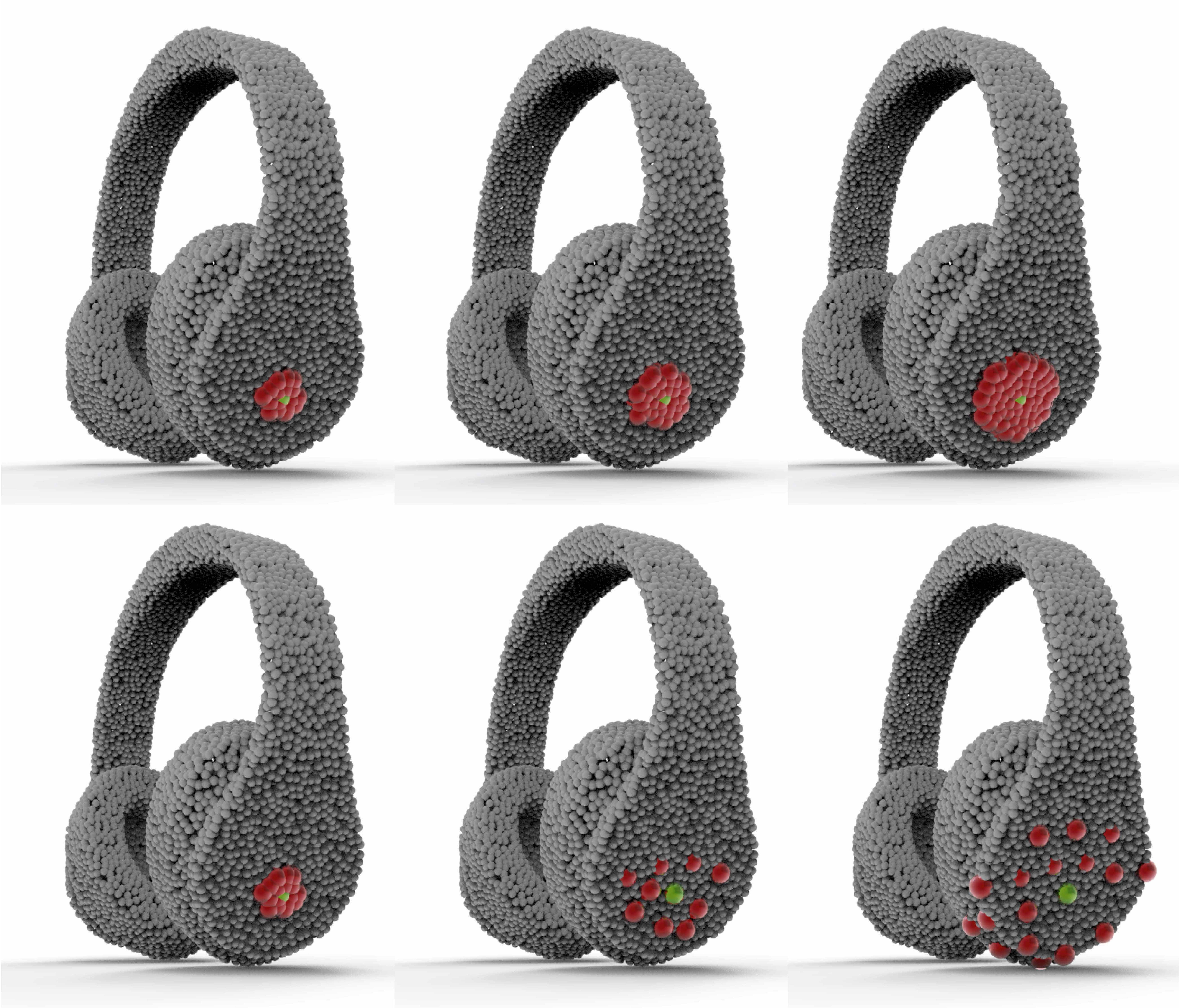}
    \centering
    \caption{Comparison of multi-scale processing (top) with multi-resolution processing (down): multi-resolution processing allows us to  process larger-scale areas while not increasing memory consumption, making it easier to elicit global context information.}
    \label{fig:multiresolution_multiscale}
\end{figure}

The \mres architecture consists in changing the way the sampling is done in the network to get a multi-resolution approach (see \reffig{fig:multiresolution_multiscale}). We provide the details only for the encoding part of the network as we keep the decoding part unchanged from \pnpp.\\
\underline{Encoding1:} \\*
$\begin{bmatrix}S(512)\rightarrow\text{rNN}(.1)\rightarrow\text{mLP}([32,32,64])\rightarrow\text{MaxP}\\S(256)\rightarrow\text{rNN}(.2)\rightarrow\text{mLP}([64,64,128])\rightarrow\text{MaxP} \\S(128)\rightarrow\text{rNN}(.4)\rightarrow\text{mLP}([64,96,128])\rightarrow\text{MaxP}\end{bmatrix}\bigoplus$\\
\underline{Encoding2:} \\
$\begin{bmatrix}S(128)\rightarrow\text{rNN}(.2)\rightarrow\text{mLP}([64,64,128])\rightarrow\text{MaxP}\\S(96)\rightarrow\text{rNN}(.4)\rightarrow\text{mLP}([128,128,256])\rightarrow\text{MaxP} \\S(64)\rightarrow\text{rNN}(.8)\rightarrow\text{mLP}([128,128,256]\rightarrow\text{MaxP}\end{bmatrix}\bigoplus$\\
\underline{Encoding3:}\\
S($1$)$\rightarrow\text{mLP}([256,512,1024])\rightarrow$ MaxP

Starting from this architecture, we add Xlink connections between each layer of each mLP to get our \mresx architecture. A Xlink connection connects two neighboring resolutions to merge information at different granularity. On each link, we use a sampling module (either downsampling or upsampling) to match the input to the target resolution. We use two alternatives for feature combination: (i) concatenation, (ii) summation. In the later case, we add an additional sLP on each Xlink to map the input feature dimension to the target. To keep this process as lean as possible, we position the SLP at the coarser resolution, i.e. before the upsampling module or after the downsampling module.

\subsubsection{LPN}
Our convPN module can be seen as a point counterpart of 2D image convolution. To do so, the convPN module replaces the MLP with its pooling layer by a sequence of SLP-Pooling modules.

To simplify the writing, we adopt the additional notations:
\begin{itemize}
    \item \textit{Sampling} block $S([s_1,s_2,..,s_n]^T)$ where we make a sampling of $s_i$ points on each resolution $i$. When only one resolution is available as input, the block $S([.,s_1,s_2,...,s_{n-1}]^T)$ will sequentially downsample the input point cloud by $s_1$, $s_2$, .. points to create the desired number of resolutions.
    \item \textit{Convolution} block $C([r_1,r_2,...,r_n]^T)$ is composed itself of three operations for each resolution $i$: neighborhood lookup to select the $r_i$NN for each points, an sLP layer of the same size as its input and a max-pooling.
    \item \textit{Transition} block $T([t_1,t_2,...,t_n]^T)$ whose main role is to change the channel dimension of the input to the one of the convolution block. An sLP of ouput dimension $t_i$ will be apply to the resolution $i$.
\end{itemize}
Residual connections are noted as *.\\
\underline{Encoding1:} \\
$S\begin{bmatrix}.\\512\\256\end{bmatrix}\rightarrow T\begin{bmatrix}32\\64\\64\end{bmatrix}\rightarrow
C^{*}\begin{bmatrix}.1\\.2\\.4\end{bmatrix}\rightarrow
T\begin{bmatrix}32\\64\\96\end{bmatrix}\rightarrow
C^{*}\begin{bmatrix}.1\\.2\\.4\end{bmatrix}\rightarrow
T\begin{bmatrix}64\\128\\128\end{bmatrix}\rightarrow
C^{*}\begin{bmatrix}.1\\.2\\.4\end{bmatrix}\rightarrow
S\begin{bmatrix}512\\256\\128\end{bmatrix}\rightarrow\bigoplus$\\
\underline{Encoding2:} \\
$S\begin{bmatrix}.\\128\\96\end{bmatrix}\rightarrow
T\begin{bmatrix}64\\128\\128\end{bmatrix}\rightarrow
C^{*}\begin{bmatrix}.2\\.4\\.8\end{bmatrix}\rightarrow
C^{*}\begin{bmatrix}.2\\.4\\.8\end{bmatrix}\rightarrow
T\begin{bmatrix}128\\256\\256\end{bmatrix}\rightarrow
C^{*}\begin{bmatrix}.2\\.4\\.8\end{bmatrix}\rightarrow
S\begin{bmatrix}128\\96\\64\end{bmatrix}\rightarrow\bigoplus$\\
\underline{Encoding3:}\\
S($1$)$\rightarrow\text{mLP}([256,512,1024])\rightarrow$ MaxP

Note here that there is no \textit{Transition} block between the first two C blocks in the Encoding2 part. This is because those two \textit{Convolution} blocks work on the same feature dimension.

We also add Xlinks inside each of the C blocks. In this architecture, the features passing through the Xlinks are combined by summation and follow the same design as for \mresx.

In the case of SLPs, using the on-the-fly re-computation of the neighborhood features tensor has a significant positive impact on both the forward and backward pass by means of a simple adjustment. Instead of applying the SLP on the neighborhood features tensor, we can first apply the SLP on the flat feature tensor and then reconstruct the neighborhood just before the max-pooling layer (Algorithm~\ref{algo:convpn_forward}). The same  can be used for the backward pass (see Algorithm~\ref{algo:convpn_backward}).

\begin{algorithm}[t!]
\DontPrintSemicolon
\KwData{Input features tensor $\mathcal{T}_f$ ($N\times R^D$), input spatial tensor $\mathcal{T}_s$ ($N\times R^3$) and indices of each point's neighborhood for lookup operation $\mathcal{L}$ ($N\times K$)}
\KwResult{Output feature tensor $\mathcal{T}_f^o$ ($N\times R^{D^{'}}$)}
\SetKwFunction{IndexLookup}{IndexLookup}
\SetKwFunction{SLP}{SLP}
\SetKwFunction{MaxPooling}{MaxPooling}
\SetKwFunction{FreeMemory}{FreeMemory}
\Begin{
\tcc*[h]{Lifting each point/feature to $R^{D^{'}}$}\;
$\mathcal{T}_{f'}\longleftarrow\SLP_f(\mathcal{T}_f)$\;
$\mathcal{T}_{s'}\longleftarrow\SLP_s(\mathcal{T}_s)$\;

\tcc*[h]{Neighborhood features $(N\times R^{D^{'}} \rightarrow N\times R^{D^{'}} \times (K+1))$}\;
$\mathcal{T}_{f'}^K\longleftarrow\IndexLookup(\mathcal{T}_{f'},\mathcal{T}_{s'},\mathcal{L})$\;

\tcc*[h]{Neighborhood pooling $(N\times R^{D^{'}}\times (K+1) \rightarrow N\times R^{D^{'}})$}\;
$\mathcal{T}_{f^{'}}^o\longleftarrow\MaxPooling(\mathcal{T}_{f^{'}}^{K})$\;
$\FreeMemory(\mathcal{T}_{s^{'}},\mathcal{T}_{f^{'}},\mathcal{T}_{f^{'}}^K)$\;
\Return $\mathcal{T}_{f^{'}}^o$
}
\caption{Low-memory grouping - Forward pass}
\label{algo:convpn_forward}
\end{algorithm}

\begin{algorithm}[t!]
\DontPrintSemicolon
\KwData{Input features tensor $\mathcal{T}_f$ ($N\times R^D$), input spatial tensor $\mathcal{T}_s$ ($N\times R^3$), gradient of the output $\mathcal{G}_{out}$ and indices of each point's neighborhood for lookup $\mathcal{L}$ ($N\times K$)}
\KwResult{Gradient of the input $\mathcal{G}_{in}$ and gradient of the weights $\mathcal{G}_{w}$}
\SetKwFunction{IndexLookup}{IndexLookup}
\SetKwFunction{InverseIndexLookup}{InverseIndexLookup}
\SetKwFunction{BackwardMaxPooling}{BackwardMaxPooling}
\SetKwFunction{BackwardSLP}{BackwardSLP}
\SetKwFunction{FreeMemory}{FreeMemory}
\Begin{
\tcc*[h]{Gradient Max Pooling $(N\times R^{D^{'}} \rightarrow N\times R^{D^{'}}\times (K+1))$}\;
$\mathcal{G}_{out}^{mp}\longleftarrow\BackwardMaxPooling(\mathcal{G}_{out})$\;
\tcc*[h]{Flattening features $(N\times R^{D^{'}}\times (K+1) \rightarrow N\times R^{D^{'}})$}\;
$\mathcal{G}_{out}^{fl}\longleftarrow\InverseIndexLookup(\mathcal{G}_{out}^{mp},\mathcal{L})$ \;
\tcc*[h]{Gradient wrt. input/weight}\;
$\mathcal{G}_{w}, \mathcal{G}_{in}\longleftarrow\BackwardSLP(\mathcal{T}_f,\mathcal{T}_s,\mathcal{G}_{out}^{fl})$\;
$\FreeMemory(\mathcal{T}_f,\mathcal{T}_s,\mathcal{G}_{out},\mathcal{G}_{out}^{mp},\mathcal{G}_{out}^{fl})$\;
\Return $(\mathcal{G}_{in},\mathcal{G}_{w})$\;
}
\caption{Low-memory grouping - Backward pass}
\label{algo:convpn_backward}
\end{algorithm}

\subsubsection{Deep LPN}
\label{appendix:deepconvpn_details}
Our deep architecture builds on LPN to design a deeper architecture. For our experiments, we double the size of the encoding part by repeating each convolution block twice. For each encoding segment, we position the sampling block after the third convolution block, so that the first half of the convolution blocks are processing a higher resolution point cloud and the other half a coarser version.

\subsection{DGCNN based architecture}
Starting from the authors' exact implementation, we swap each edge-conv layer, implemented as an MLP, by a sequence of single resolution convPN blocks. This set of convPN blocks replicates the sequence of layers used to design the MLPs in the original implementation. 

To allow the use of residual links, a transition block is placed before each edge-conv layer to match the dimension of both ends of the residual links.

\subsection{SpiderCNN based architecture}
A SpiderConv block can be seen as a bilinear operator on the input features and on a non-linear transformation of the input points. This non-linear transformation consists of changing the space where the points live in.

In the original architecture, an SLP is first applied to the transformed points to compute the points' Taylor expansion. Then, each output vector is multiplied by its corresponding feature. Finally a convolution is applied on the product. Therefore, the neighborhood features can be built \textit{on-the-fly} within the block and deleted once the outputs are obtained. We thus modify the backward pass to reconstruct the needed tensors when needed for gradient computation.

\subsection{PointCNN based architecture}
For PointCNN, we modify the $\chi$-conv operator to avoid having to store the neighborhood features tensors for the backward pass. To do so, we make several approximations from the original architecture.

We replace the first MLP used to lift the points by a sequence of convPN blocks. Thus, instead of learning a feature representation per neighbor, we retain only a global feature vector per representative point. 

We change as well the first fully connected layer used to learn the $\chi$-transformation matrix. This new layer now reconstructs the neighborhood features \textit{on-the-fly} from its inputs and deletes it from memory as soon as its output is computed. During the backward pass, the neighborhood features tensor is easily rebuilt to get the required gradients.

We implement the same trick for the convolution operator applied to the transformed features. We further augment this layer with the task of applying the $\chi$-transformation to the neighborhood features once grouped.

Finally, we place transition blocks between each $\chi$-conv layer to enable residual links.

\subsection{Implementation details}
In all our experiments , we process the dataset to have the same number of points $N$ for each sample. To reach a given number of points, input pointclouds are downsampled using the furthest point sampling (FPS) algorithm or randomly upsampled. 

We keep the exact same parameters as the original networks evaluated regarding most of parameters.

To regularize the network, we interleave a dropout layer between the last fully connected layers and parameterize it to zero 70\% of the input neurons. Finally, we add a weight decay of 5e-4 to the loss for all our experiments.

All networks are trained using the Adam optimizer to minimize the cross-entropy loss. The running average coefficients for Adam are set to $0.9$ and $0.999$ for the gradient and its square,  respectively.

\end{document}